\documentclass[journal]{IEEEtran}
\usepackage[linesnumbered,ruled]{algorithm2e}
\usepackage{graphicx,amssymb,mathrsfs,amsmath,array,color}
\usepackage{subfigure}
\usepackage{multirow}
\usepackage{booktabs}

\newcommand{\norm}[1]{\lVert#1\rVert}

\begin{document}
\title{Invariant Attribute Profiles: A Spatial-Frequency Joint Feature Extractor for Hyperspectral Image Classification}

\author{Danfeng Hong,~\IEEEmembership{Member,~IEEE,}
        Xin Wu,~\IEEEmembership{Student Member,~IEEE,}
        Pedram Ghamisi,~\IEEEmembership{Senior Member,~IEEE,}
        Jocelyn Chanussot,~\IEEEmembership{Fellow,~IEEE,}
        Naoto Yokoya,~\IEEEmembership{Member,~IEEE,}
        and~Xiao Xiang Zhu,~\IEEEmembership{Senior Member,~IEEE}

\thanks{This work is jointly supported by the German Research Foundation (DFG) under grant ZH 498/7-2, the Helmholtz Association under the framework of the Young Investigators Group SiPEO (VH-NG-1018), and the European Research Council (ERC) under the European Unions Horizon 2020 research and innovation programme (grant agreement No. ERC-2016-StG-714087, Acronym: So2Sat). The work of N. Yokoya is also supported by the Japan Society for the Promotion of Science (KAKENHI 18K18067). (\emph{Corresponding author: Xiao Xiang Zhu}).}
\thanks{D. Hong and X. Zhu are with the Remote Sensing Technology Institute (IMF), German Aerospace Center (DLR), 82234 Wessling, Germany, and Signal Processing in Earth Observation (SiPEO), Technical University of Munich (TUM), 80333 Munich, Germany. (e-mail: danfeng.hong@dlr.de; xiaoxiang.zhu@dlr.de)}
\thanks{X. Wu is with the School of Information and Electronics, Beijing Institute of Technology, 100081 Beijing, China, and Beijing Key Laboratory of Fractional Signals and Systems, 100081 Beijing, China. (e-mail: 040251522wuxin@163.com)}
\thanks{P. Ghamisi is with the Machine Learning Group, Exploration Division, Helmholtz Institute Freiberg for Resource Technology, Helmholtz-Zentrum Dresden-Rossendorf, 09599 Freiberg, Germany. (e-mail: p.ghamisi@gmail.com)}
\thanks{J. Chanussot is with the Univ. Grenoble Alpes, INRIA, CNRS, Grenoble INP, LJK, F-38000 Grenoble, France, also with the Faculty of Electrical and Computer Engineering, University of Iceland, Reykjavik 101, Iceland. (e-mail:  jocelyn@hi.is)}
\thanks{N. Yokoya is with the Geoinformatics Unit, RIKEN Center for Advanced Intelligence Project (AIP), RIKEN, 103-0027 Tokyo, Japan. (e-mail: naoto.yokoya@riken.jp)}
}

\markboth{Submission to IEEE Transactions on Geoscience and Remote Sensing,~Vol.~XX, No.~XX, ~XXXX,~2019}
{Shell \MakeLowercase{\textit{et al.}}: Rotation-Invariant Profile: A Novel Spatial-Frequency Feature Descriptor for Hyperspectral Image Classification}
\maketitle
\begin{abstract}
\textcolor{blue}{This is the pre-acceptance version, to read the final version please go to IEEE Transactions on Geoscience and Remote Sensing on IEEE Xplore.} Up to the present, an enormous number of advanced techniques have been developed to enhance and extract the spatially semantic information in hyperspectral image processing and analysis. However, locally semantic change, such as scene composition, relative position between objects, spectral variability caused by illumination, atmospheric effects, and material mixture, has been less frequently investigated in modeling spatial information. As a consequence, identifying the same materials from spatially different scenes or positions can be difficult. In this paper, we propose a solution to address this issue by locally extracting invariant features from hyperspectral imagery (HSI) in both spatial and frequency domains, using a method called invariant attribute profiles (IAPs). IAPs extract the spatial invariant features by exploiting isotropic filter banks or convolutional kernels on HSI and spatial aggregation techniques (e.g., superpixel segmentation) in the Cartesian coordinate system. Furthermore, they model invariant behaviors (e.g., shift, rotation) by the means of a continuous histogram of oriented gradients constructed in a Fourier polar coordinate. This yields a combinatorial representation of spatial-frequency invariant features with application to HSI classification. Extensive experiments conducted on three promising hyperspectral datasets (Houston2013 and Houston2018) demonstrate the superiority and effectiveness of the proposed IAP method in comparison with several state-of-the-art profile-related techniques. The codes will be available from the website: https://sites.google.com/view/danfeng-hong/data-code.
\end{abstract}
\graphicspath{{figures/}}


\section{Introduction}
\IEEEPARstart{L}{and} use and land cover (LULC) classification has been playing an increasingly vital role in the high-level image interpretation and analysis of remote sensing \cite{anderson1976land,kang2018building}. Owing to the rich spectral information, hyperspectral imagery (HSI) enables the identification and detection of the materials at a more accurate level, which has been proven to be effective for LULC-related tasks, such as HSI classification \cite{li2017hyperspectral,hang2019cascaded,gao2019spectral,yedemir2019supervised}, multi-modality data analysis \cite{hang2015matrix,hong2019cospace,hu2019comparative,hong2019learnable}, and anomaly detection \cite{li2018real,xu2018joint,wu2019approximate}. In spite of the fine spectral discrepancies in HSI, the noisy pixels, manual labeling uncertainty, and the intrinsic or extrinsic spectral variability inevitably degrade the classification performance when only the spectral profile is considered as the feature input. Fortunately, except for the spectral dimension, the two-dimensional image space can provide the extra spatial information to correct the errors in a local region by linking with different objects and robustly eliminating the effects of spectral variability \cite{hong2017unmixing,hong2018sulora,tang2018multiharmonic,hong2019augmented} between the same materials.

\begin{figure*}[!t]
	  \centering
		\subfigure{
			\includegraphics[width=0.9\textwidth]{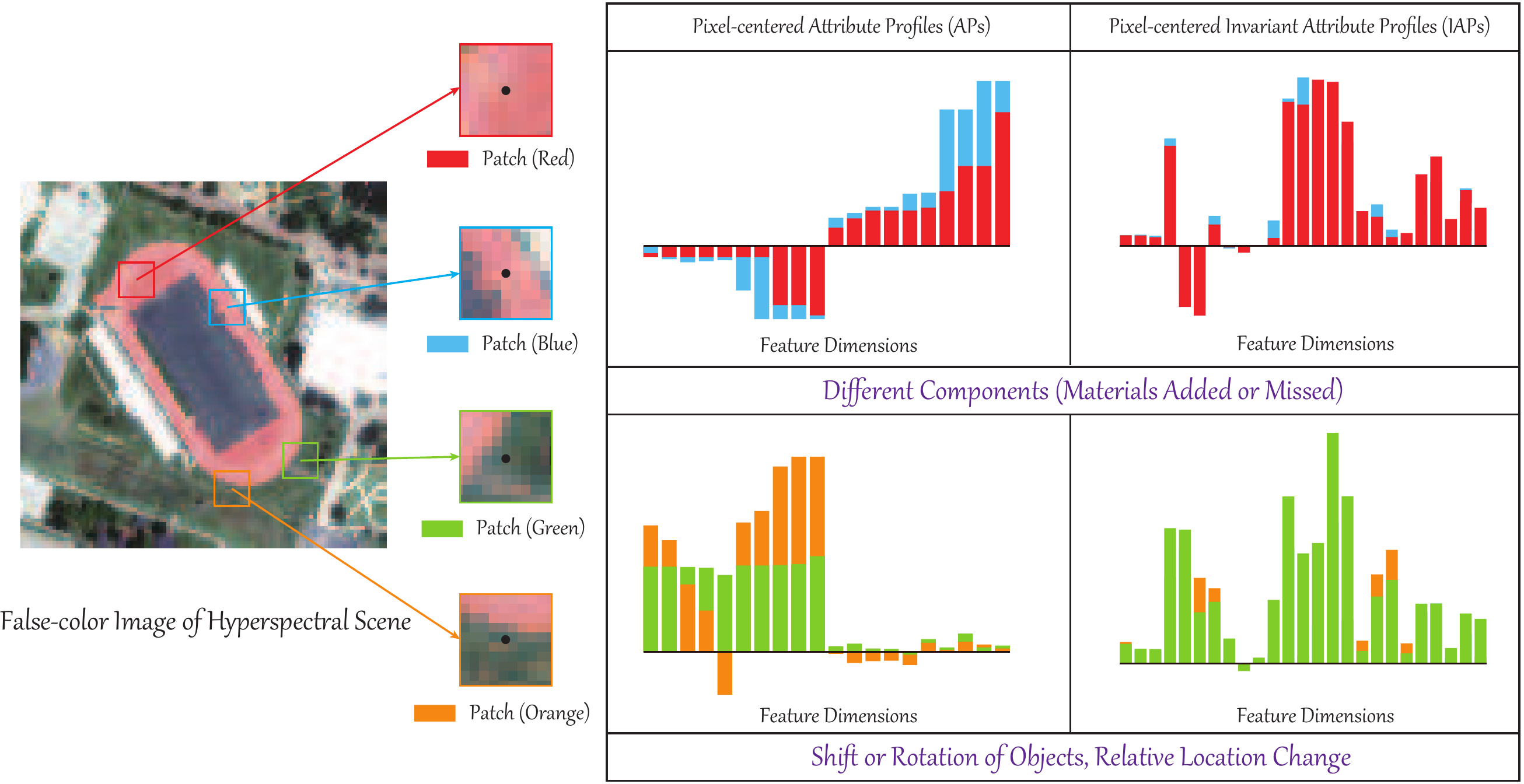}
		}
        \caption{Illustration clarifying the motivation of the proposed IAPs. Due to the difference in composition between the Red and Bule patches -- that is, the Blue patch holds more materials, there is a big difference in their APs, while the proposed IAPs are only slightly different for the two patches. Similarly, for the shift or rotation of the local scene, the IAPs are obviously more robust than the conventional APs. Given an HSI's patch, PCA is first performed on the patch to reduce its dimension to be 3. Then the APs can be extracted from the three PCs, respectively, and the final APs can be obtained by stacking all APs. The AF used in the APs is the region attribute and the code can be found in \cite{liao2017taking}. Note that the APs or IAPs are extracted based on the whole image and the cropped patches are only visual examples and the corresponding histograms are only the profile representation of the centered pixels.}
\label{fig:motivations}
\end{figure*}

In \cite{soille2003morphological}, mathematical morphology has shown its superiority in modeling and extracting the spatial information of an image related to the geometric shape and scale of different objects. Based on this concept, Pesaresi and Benediktsson \cite{pesaresi2001new} developed morphological profiles (MPs) to segment high resolution satellite imagery by applying a sequence of opening and closing operators to reconstruct or connect the targeted objects with a size-increasing structurized element (SE). The resulting morphological operator has been successfully extended and applied in the HSI classification \cite{benediktsson2005classification}, where the extended MPs (EMPs) are built on the first principal component of HSI obtained using principal component analysis (PCA) \cite{wold1987principal}. The authors of \cite{fauvel2007spectral} designed a novel strategy of jointly using SVMs and MPs for spatial-spectral hyperspectral classification. With the growing attention to MPs, a considerable volume of work related to HSI classification has frequently been reported in the literature \cite{licciardi2012linear,fauvel2013advances,hu2019comparative}. Nevertheless, the MPs' ability to extracting the diverse geometric features (e.g., textural, semantic) from hyperspectral images still remains limited, since its concept has a few limitations: for example, the shape of SEs is fixed and SEs are not able to characterize information related to the gray-level or higher-level (e.g., HSI) characteristics of the regions. More specifically, rich spectral information in HSIs makes the geometric structure of hyperspectral regions or scenes more complex, leading to difficulties in representing and extracting MPs with the SEs in an appropriate way.

To reduce the limitations of MPs, morphological attribute profiles (APs) were developed in \cite{dalla2010morphological} by applying a set of attribute filters (AFs) \cite{breen1996attribute}, connected operators utilized to process an image by considering only its connected components, to integrate the attribute components of the given image at different levels. Moreover, APs are flexible tools since they can be of different types (e.g., they can be purely geometric, or related to the spectral values of the pixels or different characteristics such as spatial relations to other connected components). APs can be viewed as a generalized extension of MPs, yet APs are advantageous over MPs due to their flexibility in capturing a variety of region-based attributes (e.g., scale, geometry, size). For example, APs allow geometrical characterization to be extracted hierarchically \cite{hong2015novel}, thereby yielding a more effective analysis in remote sensing images \cite{dalla2011classification,song2014remotely,ghamisi2015survey}.
\begin{figure*}[!t]
	  \centering
		\subfigure{
			\includegraphics[width=0.95\textwidth]{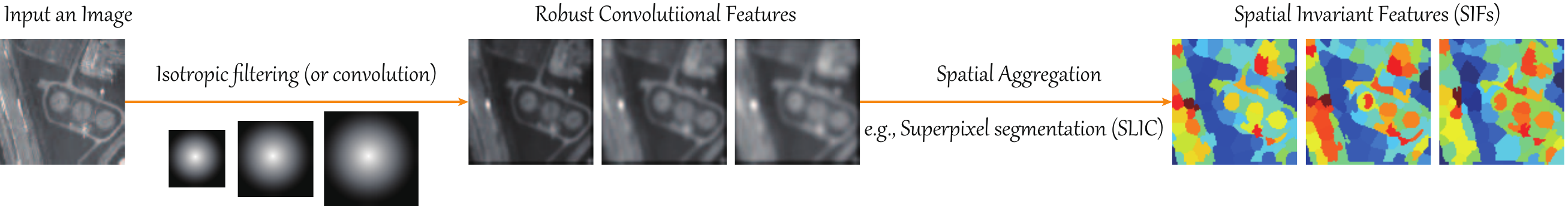}
		}
        \caption{Example illustrating the two-step extraction process of SIFs, where SLIC means simple linear iterative clustering.}
\label{fig:SIFs}
\end{figure*}

It is well known, however, that the connectivity between these defined AFs relies heavily on the geodesic reconstruction. This might lead to a problem of so-called leakage, also known as over-reconstruction \cite{liao2017taking}, where multiple regions corresponding to different objects could be merged into a single region due to improper linking. To alleviate the problem, many advanced models were proposed by using tree-based image representations to construct the AFs or APs by the means of non-redundant representations, automatic strategies, and optimization techniques \cite{xu2012morphological,falco2015spectral,demir2015histogram,cavallaro2016remote,cavallaro2017automatic}. Recently, a feasible solution to address the leakage problem was proposed in \cite{liao2016morphological} in which the AFs were partially reconstructed in order to model and extract the spatially geometrical information to a particular specification. Another representative AFs-based method to automatically and precisely extract the spatial and contextual information is extinction profiles (EPs) \cite{ghamisi2016extinction} built based on extinction filters (EFs). EPs are less sensitive to changeable image resolution, as the used EFs are determined by extrema rather than a threshold given manually in the AP. Thus, EPs have shown their effectiveness in reducing the redundant details and preserving the discriminative geometrical features, making them more useful for the classification of remote sensing data \cite{ghamisi2017lidar,fang2017extinction,xia2018random}. Similarly, the applications of APs and EPs in HSI classification are called as extended attribute profiles (EAPs) and extended extinction profiles (EEPs), respectively. Yet the APs and its variants hardly consider and investigate semantic variations. For example, two visually similar patches from a particular scene, e.g., Patch (Red) and Patch (Blue), Patch (Green) and Patch (Orange) in Fig. \ref{fig:motivations}, should share basically identical feature representation. However, in reality,
\begin{itemize}
    \item[1)] on one hand, due to the shift and rotation of pixels (objects), the extracted features inevitably suffer from substantial differences between the two similar batches: Green and Orange (see the histograms of Fig. \ref{fig:motivations} in the lower left-hand corner). More specifically, the centered pixels for the two patches stand for the same material: trees, but the locally semantic change (e.g., rotation) makes the extracted APs largely different between two patches. This might be explained by the fact that these previous APs-based methods fail to robustly merge the structure information of surrounding pixels, tending to absorb ``negative'' or ``easy-changing'' characteristics that are sensitive to a semantic change;
    \item[2)] On the other hand, despite presenting similar semantic characteristics, the slightly different components (e.g., some objects added or missed), the changed arrangement in the location and order of the objects, or spectral variability caused by illumination, topology change, atmospheric effects, and intrinsic mixing of the materials) would lead to substantial differences in the APs of the two similar patches: Red and Blue (see the histograms of Fig. \ref{fig:motivations} in the upper left-hand corner). Moreover, for Patch (Red) and Patch (Blue) that are expected to be identified as the same material from the centered pixel perspective, there is a big difference between their APs, due to the moderately semantic change (other materials involved, e.g., trees) in Patch (Blue) compared to Patch (Red). This indicates that the APs are relatively sensitive to semantic change.
\end{itemize}
Limited by the two factors described above, the resulting APs might be vulnerable and sensitive to semantic change in a local region, tending to further enlarge the negative effects on follow-up feature matching or classifier learning. Although some researchers from the remote sensing community have attempted to investigate the invariant feature representation \cite{licciardi2012retrieval} by integrating the AFs and geometric invariant moments (GIM) \cite{hu1962visual}, the poor discriminative ability of GIM limits the classification performance of the HSI to a great extent.

To overcome the challenges and pitfalls of those previously-proposed AP approaches, we propose to extract the invariant attributes (IAs) that can be robust against the semantic change in a hyperspectral scene by empowering the invariance to the AFs, yielding the proposed invariant attribute profiles (IAPs). Direct evidence is shown on the right hand side of Fig. \ref{fig:motivations}, where there are only slight perturbations between the extracted features of the two-paired patches. The IAPs consist of two parts: spatial invariant features (SIFs) and frequency invariant features (FIFs). The former are constructed in the Cartesian coordinates, where the isotropic filter banks or convolutional kernels are first exploited for HSI and the over-segmentation techniques are further used to spatially aggregate the filtered features. The latter convert discrete APs to continuous profiles by modeling the variant behaviors of pixels or objects (e.g., shift, rotation) in a Fourier polar coordinate system. More specifically, our contributions can be highlighted as follows:
\begin{itemize}
\item We propose a novel feature extractor for HSI, called invariant attribute profiles (IAPs). As the name suggests, these aim at extracting invariant feature representation by applying a sequence of well-designed AFs that are insensitive and even invariant to the change of pixels or materials in local regions.
\item The proposed approach is capable of modeling SIFs by isotropically filtering HSI in the first step. The filtered features can then be grouped into the semantically meaningful object-based representation.
\item To further improve the completeness and discrimination of the invariant features, the FIFs in this paper are also designed by extracting the continuous Fourier convolutional features in polar coordinates.
\item Two relatively new and challenging hyperspectral datasets are used to assess and compare the classification accuracies of state-of-the-art APs and our IAPs. Experimental results indicate that the classification performance of IAPs is superior to that of using the APs-based approaches, demonstrating the necessity and progressiveness of investigating and handling the issue of local semantic change in HSI classification.
\end{itemize}

The rest of this paper is organized as follows. Section II presents the methodology for extracting IAPs and the workflow for HSI classification, focusing on the design of SIFs and FIFs. In Section III, we provide the experimental results and analysis as well as brief discussion of two hyperspectral datasets from both a qualitative and quantitative perspective. Finally, our main conclusions are summarized in Section IV.

\section{Methodology}
In this section, the proposed IAPs are first introduced from two different aspects: SIFs and FIFs. The holistic workflow with the designed IAPs as the input is then developed for the HSI classification.

\subsection{Invariant Feature Extraction in the Spatial Domain} As shown in Fig. \ref{fig:motivations}, the APs in spatial domain are sensitive to various factors that can give rise to semantic change, further leading to unexpected degradation in modeling spatial information. One feasible solution for this problem is to find and extract the invariant feature representation from the image space. It is well known that isotropic filtering (or convolution) \cite{wu2018msri} is a good tool that performs robustly against shift or rotation behavior of image patches and can simultaneously eliminate various other variabilities (e.g., salt-pepper noise, missing local information) effectively. Therefore, the robust convolutional features (RCF) can be extracted from the HSI in the form of feature set $\Omega(\bullet)$
\begin{equation}
\label{eq1}
\begin{aligned}
	 \mathcal{F}_{RCF}=\Omega _{1}(\mathbf{I})=\left[ \mathcal{F}_{1}, \dots,  \mathcal{F}_{k}, \dots,  \mathcal{F}_{D}\right],
\end{aligned}
\end{equation}
where the features of the $k$-th band ($ \mathcal{F}_{k}$) can be computed by
\begin{equation}
\label{eq2}
\begin{aligned}
	  \mathcal{F}_{k}=\mathbf{I}_{k} \otimes K_{conv}.
\end{aligned}
\end{equation}
$\mathbf{I}\in \mathbb{R}^{W\times H \times D}$ is the HSI with $D$ bands by $W\times H$ pixels, where $\mathbf{I}_{k}$ denotes the $k$-th band of $\mathbf{I}$; $K_{conv}$ is defined as the convolutional kernel for isotropically aggregating the local spatial information, and the operator $\otimes$ denotes the convolutional operation.
\begin{figure*}[!t]
	  \centering
		\subfigure{
			\includegraphics[width=0.95\textwidth]{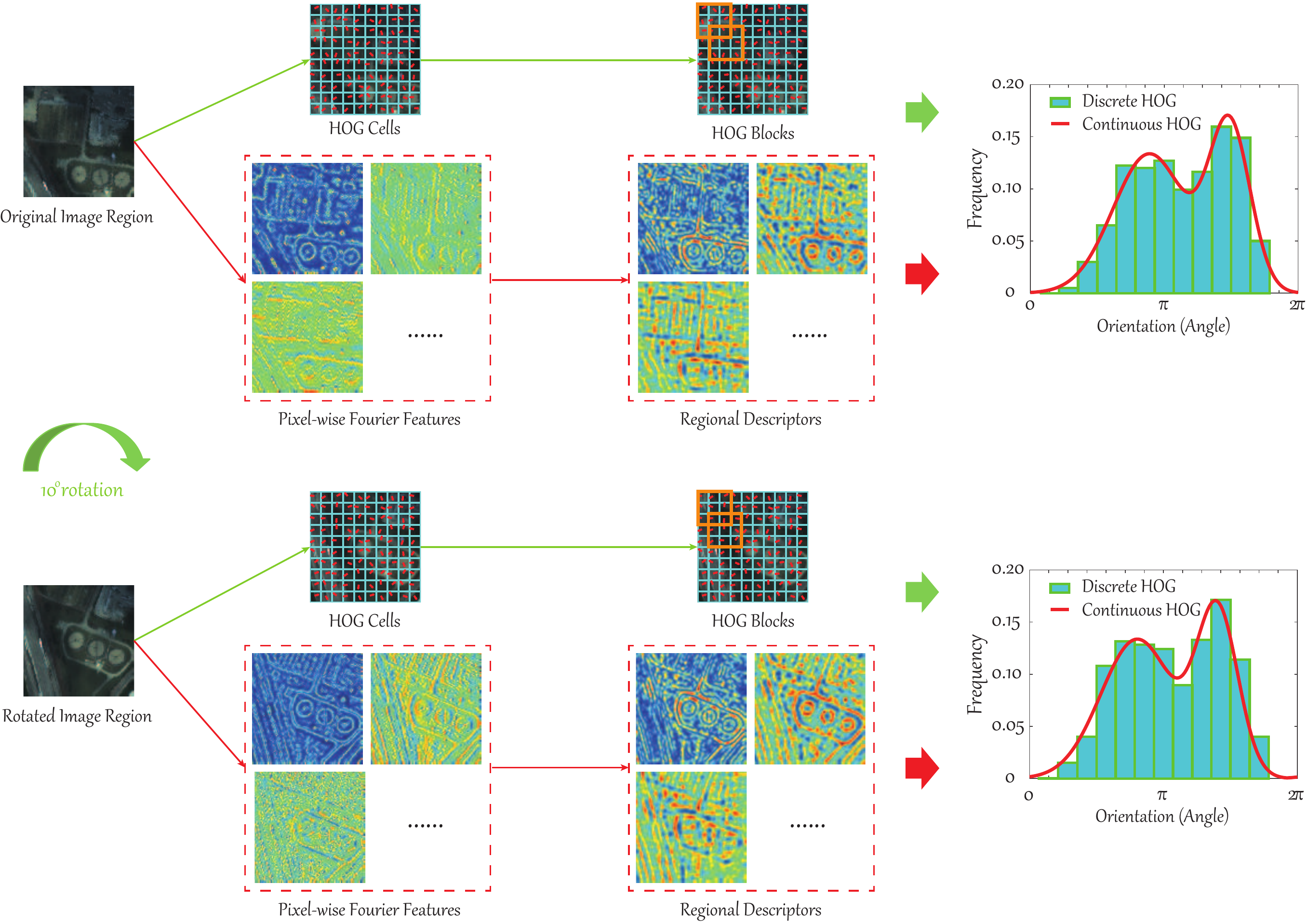}
		}
        \caption{Illustration clarifying the advantage of continuous HOG, represented by the polarized Fourier analysis from pixel-wise features to regional descriptors over the discrete HOG encoded by gradient binning (cells) and regional aggregation (blocks) in the real image regions before and after $10^{\circ}$ rotation.}
\label{fig:Ex_FIFs}
\end{figure*}

Further, we combine the filtered or convolutional features ($ \mathcal{F}_{RCF}$) with spatial aggregation (SA) techniques, such as super-pixel segmentation, to enhance the object-based semantic features, i.e., edges, shape, and the invariance of these features. In our case, we select a popular super-pixel segmentation approach: simple linear iterative clustering (SLIC) \cite{achanta2012slic}. More specifically, the SA step is implemented on the basis of RCF; thus the pixel-wise (e.g., $i$-th pixel) SIFs can be represented as
\begin{equation}
\label{eq3}
\begin{aligned}
	  \mathcal{F}_{SIFs}^{i}=\frac{1}{N_q}\sum_{j=1}^{N_q} \mathcal{F}_{RCF}^{j}, \; j\in \phi_{i,q},
\end{aligned}
\end{equation}
where $N_q$ is the number of pixels in the $q$-th super-pixel, while $\phi_{i,q}$ is defined as a pixel set ($q$-th super-pixel) including the $i$-th targeted pixel. The final SIFs are simply stacked as
\begin{equation}
\label{eq4}
\begin{aligned}
	  \mathcal{F}_{SIFs}=[ \mathcal{F}_{SIFs}^{1}, \dots,  \mathcal{F}_{SIFs}^{i}, \dots,  \mathcal{F}_{SIFs}^{N}],
\end{aligned}
\end{equation}
where $N=W\times H$ denotes the number of pixels in a given hyperspectral scene. It should be noted that in the original SLIC algorithm, the superpixels are segmented on the CIELAB color space rather than RGB space, since the differences or changes between pixels (or materials) in the CIELAB space can be perceived or captured more easily. From the perspective of feature level, CIELAB should hold a more discriminative feature representation compared to the RGB space. Naturally, the HSI is capable of better identifying the pixels due to the richer and more discriminative spectral information. However, considering the spectral redundancy in HSI, we first applied the PCA on the whole HSI to reduce the spectral dimension and simultaneously preserve the spectral information as much as possible, then performed the SLIC on the first three principal components (PCs).

The aforementioned two steps yield the SIFs of the IAPs, as illustrated in Fig. \ref{fig:SIFs}.

\subsection{Invariant Feature Extraction in the Frequency Domain}
Although SIFs are insensitive to the semantic changes in a local neighborhood, it still fails to accurately describe the shift or rotation behavior of an image (or an object) due to the quantization artifacts in a discrete coordinate system \cite{liu2014rotation}. Instead of locally estimating the discrete coordinates with the pose normalization, i.e., histogram of oriented gradients (HOG) \cite{hong2016robust}, a Fourier-based analysis technique has guaranteed the invariance of feature extraction in a polar coordinate system \cite{liu2014rotation,wu2019orsim,wu2019fourier}. More significantly, the features only extracted from the spatial domain are relatively limited in representation ability and diversity, particularly for the complex hyperspectral scene, e.g., including various spectral variabilities. Therefore, the benefits of feature extraction in the frequency domain are, on the one hand, to enrich the diversity of the features, thereby further improving the performance of HSI classification; and, on the other hand, to be capable of effectively modeling the invariant behaviors by the the means of continuous signal representation, yielding robust feature extraction against a variety of semantic change in HSI.

For example, the rotation behavior of an object or local image patch can not be well modeled by a discrete histogram, but can be explained by circular shift by continuous and smooth signal representation. For this reason, the continuous Fourier transformation has been proven to be an effective tool to model the rotation behavior with any angles,  e.g., either integer or non-integer, more accurately \cite{liu2014rotation}. Fig. \ref{fig:Ex_FIFs} illustrates the two strategies for extracting feature descriptors for the traditional discrete HOG and continuous Fourier-based HOG. In detail, the differences from the pixel-wise feature design to the spatially regional feature aggregation are clarified by the two approaches, respectively. Another can be seen in Fig. \ref{fig:Ex_FIFs}, where there is a relatively obvious change between two image regions or patches in the form of a discrete histogram (e.g., HOG), due to the $10^{\circ}$ rotation behavior. Conversely, the Fourier-based HOG descriptor is robust to the big semantic gap when the image scene changes significantly, which only yields a slight shift along the horizontal coordinate and basically keeps the feature shape unchanged. In the following, we will detail the procedure for extracting the Fourier-based continuous HOG from the pixel-wise features to the region-based representation.

\subsubsection{Analysis of Rotation-invariance} As the name suggests, rotation-invariance refers to the extracted features $f(x,y)$ for a given pixel located in the $(x,y)$ of the image remaining the same or unchanged when the image rotates with a $g^{o}$ angle. Similar to \cite{wu2019fourier}, the rotation behavior can be formulated as
\begin{equation}
\label{eq5}
\begin{aligned}
	  gf(x,y)=f(x,y),\;\; or\;\; h(\mathbf{I}(x,y)\circ \mathbf{T}_{g})=h(\mathbf{I}(x,y)),
\end{aligned}
\end{equation}
where $h(\mathbf{I})$ is abstracted as a feature extractor from the input image $\mathbf{I}$, and $\mathbf{T}_{g}$ represents an operation of coordinate transformation. For those pixels that do not change the locations, the invariant features can be simply deduced by Eq. (\ref{eq5}) as $ gf(x,y)=f(x,y)$. For other pixels whose locations are changed, the coordinate transformation $\mathbf{T}_{g}$ has to be considered to formulate the rotation-invariance \cite{wu2019fourier} as
\begin{equation}
\label{eq6}
\begin{aligned}
	  gf(x,y)=f(\mathbf{T}_{g}(x,y))=f(x,y)\circ\mathbf{T}_{g}.
\end{aligned}
\end{equation}
We then have the equivalent condition $gf=f\circ\mathbf{T}_{g}$, which has been proven and widely used in many works \cite{reisert2008equivariant,wang2009rotational,vedaldi2011learning}.

\subsubsection{Pixel-wise Fourier Features} Let a 2-D location of any one pixel in a given image $\mathbf{I}$ be $(x,y)$ in Cartesian coordinates or $(\norm{\mathbf{D}(x,y)},\theta(\mathbf{D}(x,y)))$ in polar coordinates, where  $\norm{\mathbf{D}(x,y)}$ and $\theta(\mathbf{D}(x,y))$ are defined as the magnitude and the phase information of a complex number: $\mathbf{D}(x,y)=dx+dyi$, and $dx$ and $dy$ are denoted as the horizontal and vertical gradients, respectively, in the location of $(x,y)$ of $\mathbf{I}$ in  Cartesian coordinates. Thus, given an input pixel $p$, its $m$-th order Fourier representation, denoted as $\mathbf{F}_{m}(x,y)$, can be obtained by
\begin{equation}
\label{eq7}
\begin{aligned}
	  \mathbf{F}_{m}(x,y)&=\frac{1}{2\pi}\int_{0}^{2\pi}h(\varphi)e^{-im\varphi} \\
	  &= \norm{\mathbf{D}(x,y)}e^{-im\theta(\mathbf{D}(x,y))},
\end{aligned}
\end{equation}
where $h(\varphi)$ is an orientation distribution function with respect to the angle $\varphi$ in the pixel $p$. This function can be given  by an impulse response function $\delta$ with integral $\norm{\mathbf{D}(x,y)}$:
\begin{equation}
\label{eq8}
\begin{aligned}
	  h(\varphi):=\norm{\mathbf{D}(x,y)}\delta(\varphi-\theta(\mathbf{D}(x,y))).
\end{aligned}
\end{equation}
Unlike the Fourier transformation in Cartesian coordinates, the polarized Fourier transformation has been proven to be effective for separating the angular information ($e^{-im\theta(\mathbf{D}(x,y))}$) and radial basis from the Fourier representations \cite{liu2014rotation}. As shown in Eq. (\ref{eq7}), the rotated angle in Cartesian coordinates can be equivalently explained by a shift behavior in the polar representation.
\begin{figure}[!t]
	  \centering
		\subfigure{
			\includegraphics[width=0.45\textwidth]{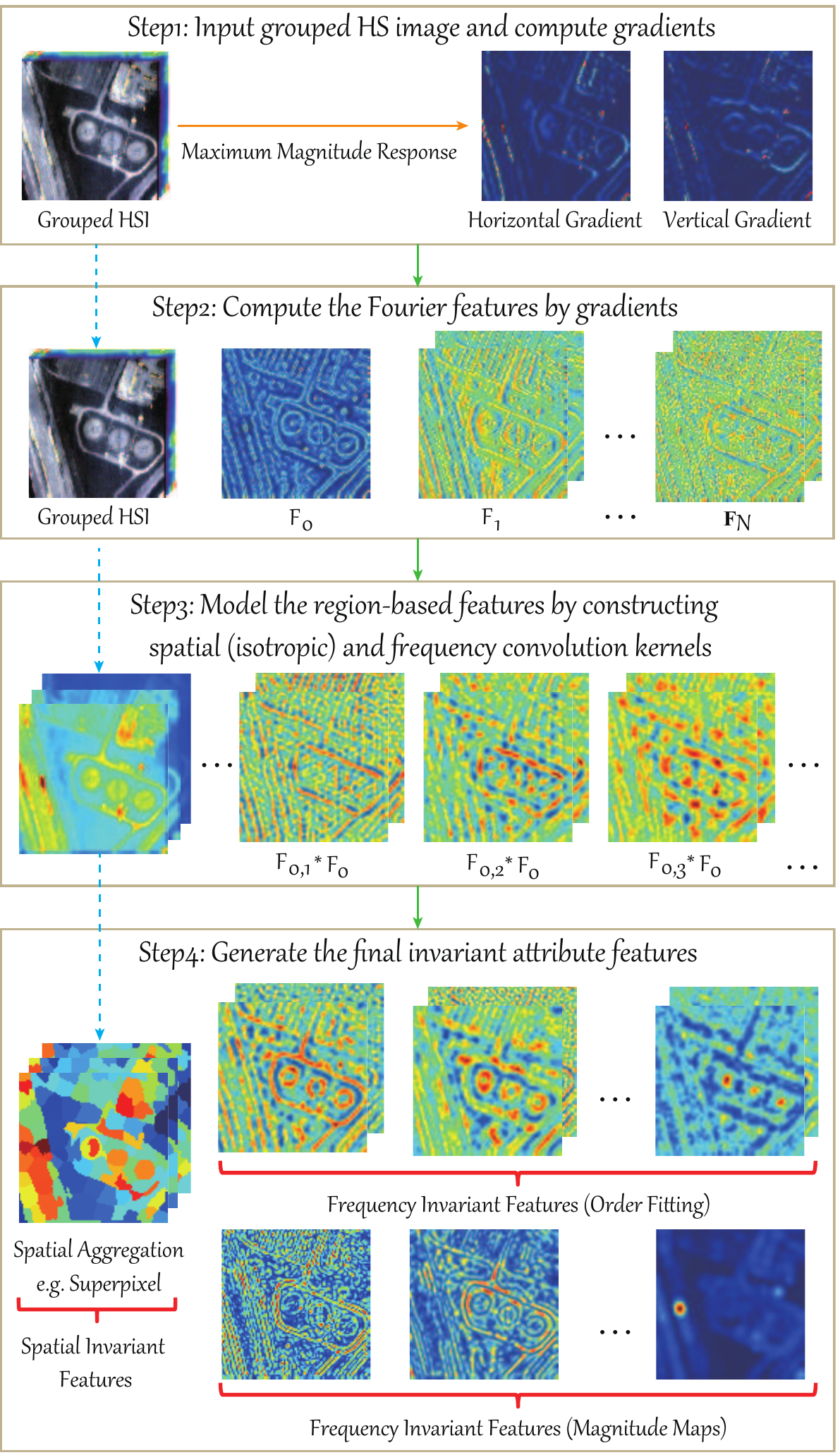}
		}
        \caption{A step-wise workflow to extract the proposed IAPs that consist of spatial invariant features (SIFs) and frequency invariant features (FIFs) of order fitting and magnitude maps.}
\label{fig:IAPs}
\end{figure}
\begin{figure*}[!t]
	  \centering
		\subfigure{
			\includegraphics[width=0.95\textwidth]{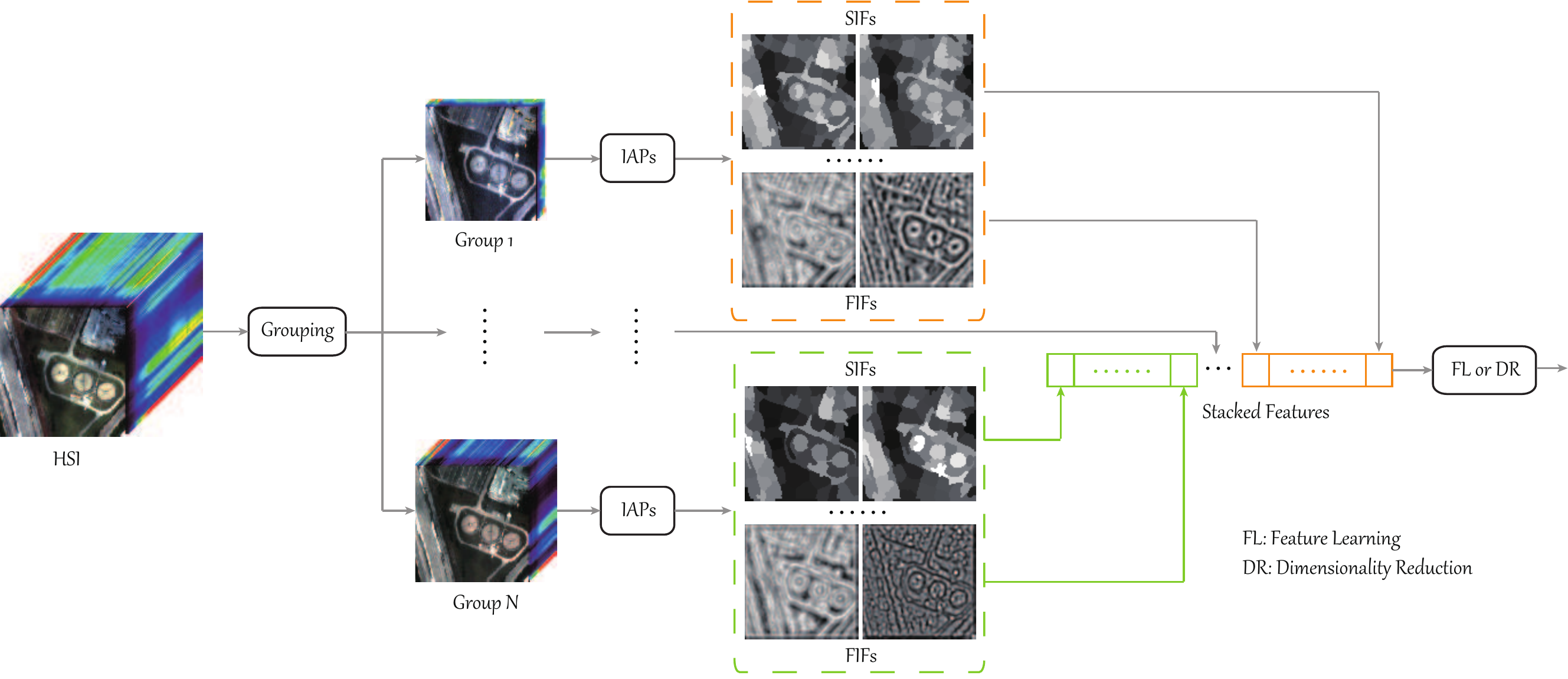}
		}
        \caption{The proposed IAPs-based classification framework.}
\label{fig:Classification}
\end{figure*}

As shown in Eq. (\ref{eq7}), one natural and intuitive way to eliminate the effects of the image rotation in feature extraction is to fit the angular information. It is, however, hardly possible to directly estimate the phase information ($\theta(\mathbf{D}(x,y))$), due to its complexity and uncertainty. Alternatively, we may enforce the Fourier order $m$ to be zero, thereby achieving the same goal for removing the phase information. When a $g^{o}$ rotation is applied on $\mathbf{F}_{m}(x,y)$, accordingly to the mathematical property of Fourier transformation in Polar coordinates \cite{liu2014rotation}, we then deduce the rotated Fourier representation $g\mathbf{F}_{m}(x,y)$ as
\begin{equation}
\label{eq9}
\begin{aligned}
	  g\mathbf{F}_{m}(x,y)&=\left[\norm{\mathbf{D}(x,y)}e^{-im(\theta(\mathbf{D}(x,y))+g^{o})}\right]\circ\mathbf{T}_{g}\\
	  &=e^{-img^{o}}\left[\mathbf{F}_{m}(x,y)\circ\mathbf{T}_{g}\right].
\end{aligned}
\end{equation}
Owing to the self-steerability of the Fourier basis under polar coordinates \cite{jacovitti2000multiresolution} -- that is, such a basis can be self-steered to any orientation -- we can construct a sequence of Fourier base with self-adaptive rotation angles. According to the special property, the rotated Fourier representation in Eq. (\ref{eq9}) can be multiplied or convoluted by another Fourier basis with the same rotation behavior (e.g., a $g^{o}$ angle):
\begin{equation}
\label{eq10}
\begin{aligned}
	  g(&\mathbf{F}_{m}(x,y)*\mathbf{F}_{m^{'}}(x,y))\\
	  &=e^{-i(m+m^{'})g^{o}}\left[\mathbf{F}_{m}(x,y)*\mathbf{F}_{m^{'}}(x,y)\right]\circ\mathbf{T}_{g},\\
	  g(&\mathbf{F}_{m}(x,y)\cdot\mathbf{F}_{m^{'}}(x,y))\\
	  &=e^{-i(m+m^{'})g^{o}}\left[\mathbf{F}_{m}(x,y)\cdot\mathbf{F}_{m^{'}}(x,y)\right]\circ\mathbf{T}_{g},
\end{aligned}
\end{equation}
where $\ast$ and $\cdot$ operators denote the convolution and multiplication behaviors, respectively. In order to make the extracted features rotation-invariant, we have to meet the condition of Eq. (\ref{eq6}): that is, $gf(x,y)=f(x,y)\circ\mathbf{T}_{g}$. It is, therefore, natural to have a solution of rotation-invariance as long as $m+m^{'}=0$ is satisfied.
Note that the convolutional representation of Eq. (\ref{eq10}) is regarded as the final rotation-invariant output.

Supported by the above analysis and derivation of rotation-invariance in theory, we further detail the procedures to extract the rotation-invariant features from the images in practice. This process consists of three parts, as follows:
\begin{itemize}
    \item \textit{Part 1: magnitude.} By applying the polarized Fourier transformation on each pixel of the input image, we can obtain the $m$ magnitude features corresponding to $m$ different Fourier orders, which can be formulated  for each pixel as $\mathcal{F}^{1}_{m}(x,y)=\norm{\mathbf{D}_{m}(x,y)}, \; m=0,1,\dots,m$.
    \item \textit{Part 2: absolute rotation-invariant features.} The rotation behavior can be compensated by completely removing the phase information by using Eq. (\ref{eq10}) and arranging $m+m^{'}=0$. Then, the features of part 2 can be represented as $\mathcal{F}^{2}_{m}(x,y)=\mathbf{F}_{m}(x,y)*\mathbf{F}_{m^{'}(x,y)},$ and $m=-m^{'}$.
    \item \textit{Part 3: relative rotation-invariant features.} To reduce the loss of rich phase information, the relative rotation-invariant features are developed by coupling the adjunct convolutional Fourier representations obtained with two neighbouring convolutional kernel-radii \cite{giannakis1989signal}. This process can be performed by
    \begin{equation}
    \label{eq11}
    \begin{aligned}
	      &\mathcal{F}^{3}_{m}(x,y)=\\
	      &\frac{(\mathbf{F}_{m}(x,y)*\mathbf{F}_{m^{'},r1}(x,y))\overline{(\mathbf{F}_{m}(x,y)*\mathbf{F}_{m^{'},r2}(x,y))}}{\sqrt{\norm{(\mathbf{F}_{m}(x,y)*\mathbf{F}_{m^{'},r1}(x,y))\overline{(\mathbf{F}_{m}(x,y)*\mathbf{F}_{m^{'},r2}(x,y))}}}},
    \end{aligned}
    \end{equation}
    which is subject to $m\neq m^{'}$. The terms $r_1$ and $r_2$ are the radii of two different convolutional kernels, and the symbol $\overline{\mathbf{F}}$ is defined as the complex conjugate.
\end{itemize}

Combined with the three parts, the final pixel-wise Fourier features (PWFF) can be written as
\begin{equation}
\label{eq12}
\begin{aligned}
	      \mathcal{F}_{PWFF}&(x,y)=\\
	      &[\mathcal{F}_{0}^{1}(x,y),\dots,\mathcal{F}_{m}^{1}(x,y),\dots,\mathcal{F}_{0}^{2}(x,y),\\
	      &\dots,\mathcal{F}_{m}^{2}(x,y),\dots,\mathcal{F}_{0}^{3}(x,y),\dots,\mathcal{F}_{m}^{3}(x,y)].
\end{aligned}
\end{equation}
By collecting all PWFF, we then have
\begin{equation}
\label{eq13}
\begin{aligned}
	      \mathcal{F}_{PWFF}=\{\{\mathcal{F}_{PWFF}(x,y)\}_{x=1}^{W}\}_{y=1}^{H}.
\end{aligned}
\end{equation}

\subsubsection{Regional Descriptors} Analogous to the HOG blocks, whcih aim to describe object-based contextual information, we aggregate the PWFF into region-based descriptors with the use of isotropically triangular convolutional kernels. In our case, multi-scaled convolutional kernels are used to capture the semantic information of different receptive fields. The scaling setting will be discussed in the experimental section. Finally, the resulting FIFs are
\begin{equation}
\label{eq14}
\begin{aligned}
	      \mathcal{F}_{FIFs}=\left[\mathcal{F}_{PWFF}^{C_{1}},\dots, \mathcal{F}_{PWFF}^{C_{j}},\dots\right],
\end{aligned}
\end{equation}
where $\mathcal{F}_{PWFF}^{C_{j}}$ denotes the regional descriptors with the $j$-th convolutional kernel.

\subsection{Invariant Attribute Profiles (IAPs)}
For our proposed IAPs, the fusion strategy of SIFs and FIFs is nothing but a stacking operation. Despite its simplicity, this fusion strategy has been widely and successfully applied in various feature extraction tasks. Fig. \ref{fig:IAPs} illustrates the step-wise workflow of extracting SIFs and FIFs, which can be delineated more specifically as follows.

\begin{itemize}
    \item \textit{Step 1: Group the HSI and compute its gradients.} Due to the redundancy of HSI, some adjunct bands might share similar invariance. To address that circumstance, we first group the HSI, e.g., using clustering techniques ($k$-means algorithm in our case), then compute the horizontal and vertical gradients for each group by means of the maximum magnitude response (see the first step in Fig. \ref{fig:IAPs}). Note that the final classification performance is sensitive to the number of the grouped HSIs, that is, excessively large or small ones would make the information redundant or coupled across bands. Therefore, a proper parameter setting is needed. In our case, the number of the grouped HSIs can be effectively determined by cross-validation on the labeled training set.
    \item \textit{Step 2: Extract polarized Fourier features by gradients.} The Fourier complex form can be generated by gradients (see the subsection entitled \textit{Pixel-wise Fourier Features}). Thus, the polarized Fourier representation given in Eq. (\ref{eq7}) can be obtained by applying Fourier transform to the complex number.
    \item \textit{Step 3: Construct the regional descriptors on both spatial and frequency domains.} We model the region-based representation by using isotropically spatial filters and Fourier convolutional kernels.
    \item \textit{Step 4: Generate the proposed IAPs.} The spatial aggregation technique, i.e., superpixel segmentation, is used to generate the SIFs. In the frequency domain, there are two parts in the FIFs: magnitude and order fitting. The latter consists of absolute rotation-invariant features and relative rotation-invariant features. In that way, the IAPs can be abstracted as $\mathcal{F}_{IAPs}=\left[\mathcal{F}_{SIFs},\mathcal{F}_{FIFs}\right]$.
\end{itemize}

\subsection{IAPs-based HSI Classification Framework}
With the proposed IAPs, we intend to develop an automatic HSI classification system. For this purpose, an IAPs-based HSI classification framework is designed. As shown in Fig. \ref{fig:Classification}, the framework is mainly composed of HSI grouping, IAPs extraction, feature stacking, and feature learning (FL) (or dimensionality reduction (DR)) \cite{hong2017learning}. For more details, the HSI is grouped in a band-wise way, while the feature learning step is simply and effectively conducted by PCA, in our case. In the experiments below, we found that the FL or refinement step plays a significant role in improving the classification performance. This could be reasonably explained by the redundant and noisy concentrated features.

\subsection{Feasibility and Effectiveness Analysis of the Proposed IAPs for HSI Classification}
Different from pixel-wise semantic labeling of natural images, an effective and accurate HSI classification algorithm usually depends on jointly modeling spatial and spectral information. Owing to the effective spatial structure (or pattern) modeling, the MPs (or APs) and their variants have been widely and successfully applied for HSI classification. With this motivation, the proposed IAPs similarly consider the spatially structural information in the form of semantic patches or objects with spatial aggregation strategy (e.g., superpixel), yielding SIFs. In \cite{dollar2014fast}, the SIFs can be regarded as the low-level shift-invariant feature representation, which has been theoretically proven to be effective for robustly addressing variability within the same class. On the other hand, the IAPs model the irregular textural information from the frequency domain, yielding FIFs, in order to further improve the feature discrimination. Similarly, there is also a good theoretical support for the extracted FIFs in \cite{liu2014rotation,wu2019orsim}, demonstrating its effectiveness in extracting the invariant features from optical remote sensing images for classification and detection tasks. The compact combination of the two features with their invariant attributes makes it possible for the proposed IAPs to classify the HSI more robustly and accurately. More notably, unlike those previously-proposed MPs and APs that are only performed on first few components obtained by PCA, our IAPs are not only capable of extracting spatial semantic features with spatial aggregation techniques and convolutional (or filtering) operators, but also fully considering the spectral information by the means of grouping strategy, maximum magnitude response, and FL or DR after feature extraction instead of DR before feature extraction (e.g., using PCA).
\begin{table}[!t]
\centering
\caption{Scene categories of the Pavia University dataset， with the number of training and test samples shown for each class.}
\begin{tabular}{cccc}
\toprule[1.5pt]
Class No.&Class Name&Training&Test\\
\hline \hline 1&Asphalt&548&6631\\
 2&Meadows&540&18649\\
 3&Gravel&392&2099\\
 4&Trees&524&3064\\
 5&Metal Sheets&265&1345\\
 6&Bare Soil&532&5029\\
 7&Bitumen&375&1330\\
 8&Bricks&514&3682\\
 9&Shadows&231&947\\
\hline \hline &Total&3921&42776\\
\bottomrule[1.5pt]
\end{tabular}
\label{Table:Pavia}
\end{table}
\begin{table}[!t]
\centering
\caption{Scene categories of the Houston2013 dataset， with the number of training and test samples shown for each class.}
\begin{tabular}{cccc}
\toprule[1.5pt]
Class No.&Class Name&Training&Test\\
\hline \hline 1&Healthy Grass&198&1053\\
 2&Stressed Grass&190&1064\\
 3&Synthetic Grass&192&505\\
 4&Tree&188&1056\\
 5&Soil&186&1056\\
 6&Water&182&143\\
 7&Residential&196&1072\\
 8&Commercial&191&1053\\
 9&Road&193&1059\\
 10&Highway&191&1036\\
 11&Railway&181&1054\\
 12&Parking Lot1&192&1041\\
 13&Parking Lot2&184&285\\
 14&Tennis Court&181&247\\
 15&Running Track&187&473\\
\hline \hline &Total&2832&12197\\
\bottomrule[1.5pt]
\end{tabular}
\label{Table:H2013}
\end{table}
\begin{table}
\centering
\caption{Scene categories of the Houston2018 dataset, with the number of training and test samples shown for each class.}
\begin{tabular}{cccc}
\toprule[1.5pt]
 Class No.&Class Name&Training&Test\\
\hline \hline 1&Healthy Grass&500&9299\\
 2&Stressed Grass&500&32002\\
 3&Artificial Turf&68&616\\
 4&Evergreen Trees&500&13088\\
 5&Deciduous Trees&500&4548\\
 6&Bare Earth&451&4065\\
 7&Water&26&240\\
 8&Residential Buildings&500&39262\\
 9&Non-Residential Buildings&500&223184\\
 10&Roads&500&45310\\
 11&Sidewalks&500&33502\\
 12&Crosswalks&151&1365\\
 13&Major Thoroughfares&500&45858\\
 14&Highways&500&9349\\
 15&Railways&500&6437\\
 16&Paved Parking Lots&500&10975\\
 17&Unpaved Parking Lots&14&135\\
 18&Cars&500&6078\\
 19&Trains&500&4865\\
 20&Stadium Seats&500&6324\\
\hline \hline &Total&8210&496502\\
\bottomrule[1.5pt]
\end{tabular}
\label{Table:H2018}
\end{table}

\section{Experiments}
\subsection{Data Description}
We quantitatively and qualitatively evaluate the algorithm performance on three representative and promising HSI datasets: Pavia University, Houston2013, and Houston2018, in the form of image classification. These two datasets are briefly introduced as follows.

\subsubsection{Pavia University Dataset} The hyperspectral scene was acquired by the ROSIS sensor over the campus of Pavia University, Paiva, Italy. The image consists of $610\times 340$ pixels at a ground sampling distance (GSD) of $1.3m$ with 103 spectral bands in the range of $430nm$ to $860nm$. In this dataset, nine main categories are investigated for land cover classification task. The number of training and test samples are specifically listed in Table I, while the corresponding sample distribution is given in Fig. \ref{fig:CM_Pavia}.

\subsubsection{Houston2013 Dataset} The ITRES CASI-1500 sensor was used to acquire the data over the campus of University of Houston and its surrounding areas, in Houston, Texas, USA. This dataset was provided for the 2013 IEEE GRSS data fusion contest, and is composed of $349\times 1905$ pixels with $144$ spectral channels ranging from $364nm$ to $1046nm$ at a spectral sampling of $10nm$. There are 15 challenging classes in the form of LULC in the scene. Table \ref{Table:H2013} lists the scene categories and the number of training and test samples used in the classification task, while Fig. \ref{fig:CM_Houston2013} visualizes the false-color image of the hyperspectral scene and the sample distribution of the training and test set.

\subsubsection{Houston2018 Dataset} The dataset is an airborne multi-modal data product, where the hyperspectral data was acquired by the same IRTES CASI-1500 sensor at a GSD of $1m$. This data has become available from the 2018 IEEE GRSS data fusion contest and its size is $601\times 2384$ with 50 spectral bands sampling the wavelength of between $380nm$ to $1050nm$ at intervals of $10nm$. The specific training and test information for the data is detailed in Table \ref{Table:H2018}, and a false-color image and the locations of labeled samples of training and test set are also given in Fig. \ref{fig:CM_Houston2018}.

In the second dataset, we artificially chose a number of challenging pixels to act as the training samples with multiple different experiments on the given ground truth (see \cite{marpu2012classification} for more details). The benefits of the strategy to separate the training and test sets are two-fold. On one hand, unlike random selection for training and test set that usually yields a very high performance, such a challenging setting may help us assess the potential and performance of extracted features more effectively. On the other hand, using the fixed training and test samples also contributes to making a consistent fair comparison by reproducing the experimental results using different methods.
\begin{figure*}[!t]
	  \centering
		\subfigure{
			\includegraphics[width=1\textwidth]{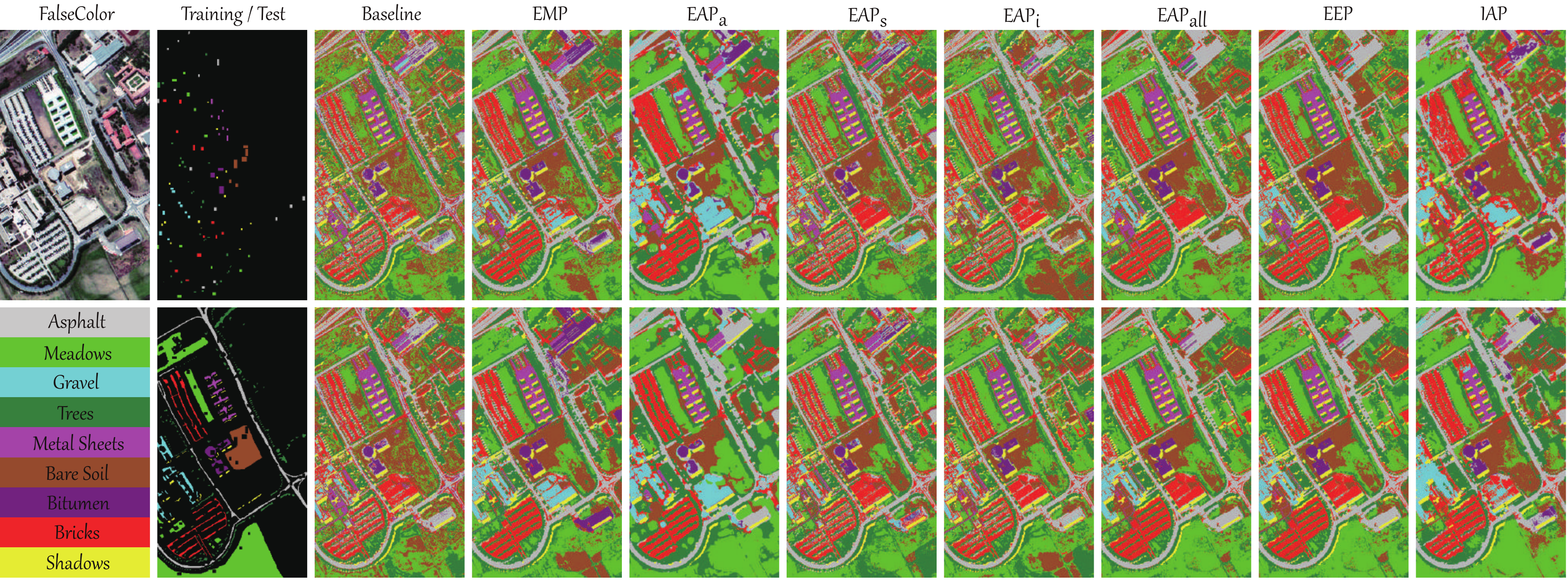}
		}
        \caption{Scene categories and visualization of false-color image, the distribution of training and test samples, and classification maps of different compared algorithms using two classifiers (top: NN, bottom: RF) on the Pavia University dataset.}
\label{fig:CM_Pavia}
\end{figure*}
\begin{table*}[!t]
\centering
\caption{Quantitative classification performance for eight different algorithms on the Pavia university dataset.}
\resizebox{\textwidth}{!}{
\begin{tabular}{c|cc|cc|cc|cc|cc|cc|cc|cc}
\toprule[1.5pt]Methods&\multicolumn{2}{c|}{Baseline (103) }&\multicolumn{2}{c|}{EMP (63) }&\multicolumn{2}{c|}{EAP$_a$ (63) }&\multicolumn{2}{c|}{EAP$_s$ (63) }&\multicolumn{2}{c|}{EAP$_i$ (63) }&\multicolumn{2}{c|}{EAP$_{all}$ (189) }&\multicolumn{2}{c|}{EEP (213)}&\multicolumn{2}{c}{IAP (30) }\\
\hline Classifier& NN & RF & NN & RF & NN & RF & NN & RF & NN & RF & NN & RF & NN & RF & NN & RF\\
\hline OA&71.85&71.50&78.22&81.19&79.87&84.35&78.24&78.64&69.24&75.30&83.68&89.65&85.84&91.96&92.43&\bf95.78\\
              AA&81.15&82.13&88.05&89.40&83.68&86.96&84.37&83.61&82.47&84.03&89.45&91.47&87.38&89.73&91.94&\bf93.15\\
              $\kappa$&0.6501&0.6499&0.7279&0.7602&0.7414&0.7956&0.7259&0.7286&0.6275&0.6894&0.7914&0.8629&0.8156&0.8920&0.9005&\bf0.9441\\
\hline Class1&73.17&80.86&90.86&93.55&74.97&82.51&82.34&87.45&88.19&91.66&91.98&96.15&77.03&88.99&87.65&\bf96.40\\
              Class2&61.32&55.64&63.69&71.23&78.64&85.06&71.22&73.29&48.64&64.13&75.97&90.91&87.31&\bf99.08&94.88&98.01\\
              Class3&60.17&51.74&72.84&77.13&70.65&76.46&60.03&52.36&58.55&57.60&72.56&\bf78.99&70.08&59.31&75.46&69.56\\
              Class4&97.39&98.73&98.20&98.83&99.35&98.43&96.34&99.48&96.83&98.20&99.15&97.68&\bf99.51&99.18&91.19&97.45\\
              Class5&99.26&99.11&99.48&99.70&98.07&\bf100.00&99.41&98.96&99.03&99.26&99.48&99.70&98.59&\bf100.00&99.48&96.65\\
              Class6&73.35&79.16&76.38&66.16&62.99&63.07&78.86&65.22&76.42&60.49&75.46&61.64&76.44&69.80&95.01&\bf95.82\\
              Class7&84.96&84.44&97.74&99.62&79.10&82.78&87.67&89.85&95.64&95.04&\bf99.92&\bf99.92&87.59&95.56&94.66&90.08\\
              Class8&85.36&91.53&99.02&\bf99.62&97.66&97.64&84.44&89.98&86.45&91.93&98.70&99.40&97.91&98.51&91.01&98.67\\
              Class9&95.35&97.99&94.19&98.73&91.66&96.73&\bf99.05&95.88&92.50&97.99&91.87&98.84&91.97&97.15&98.1&95.67\\
\bottomrule[1.5pt]
\end{tabular}}
\label{tab:pavia}\\
\footnotesize{$^{*}$ The best is shown in bold. The feature dimensions are given in the parentheses after the method's name.}
\end{table*}
\subsection{Experimental Setup}
\subsubsection{Evaluation Metrics} Pixel-wise classification is explored as a potential application for quantitatively evaluating the performance of these feature extraction algorithms. Three commonly-used criteria, \textit{Overall Accuracy (OA)}, \textit{Average Accuracy (AA)}, and \textit{Kappa Coefficient ($\kappa$)}, are adopted to intuitively quantify the experimental results using two simple classifiers: nearest neighbor (NN) and random forest (RF). The two classifiers have been widely used in many works related to HSI classification \cite{hong2018joint,hong2019learning}. Please note that the reason for selecting the two classifiers in our case is to emphasize the performance gain from the features rather than from the complex and advanced classifiers.

\subsubsection{State-of-the-Art Comparison in Related Works} The MP-based or AP-based methods we investigate in this paper are obviously categorized into unsupervised feature extraction. Therefore, the spatial-spectral features are extracted from the whole HSI. To demonstrate the effectiveness and superiority of IAPs, the baseline (original spectral features (OSF)) and other six advanced MPs or APs are compared with the proposed method. Specifically, they are EMPs, EAPs$_{a}$, EAPs$_{s}$, EAPs$_{i}$, EAPs$_{all}$, and EEPs, where $a$, $s$, $i$, and $all$ denote the attributes for the area of the regions, the standard deviation in the regions, moment of inertia, and the stack of the previous three attributes, respectively.

\subsubsection{Algorithm Configuration}
As a matter of routine, PCA was performed on the original HSI to obtain several PCs (the first three PCs in our case) before applying opening and closing operations in MPs or thinning and thickening ones in APs. Moreover, the parameters for MPs and each attribute of APs are set by following the studies in \cite{liao2017taking}, specifically:
\begin{itemize}
    \item MPs: the scales are 10, that is, there were ten openings and closings;
    \item AP$_{a}$: $\lambda_{a}=[100, 500, 1000, 2000, 3000,\dots,7000, 8000]$;
    \item AP$_{s}$: $\lambda_{s}=[0.1,0.5,1,2,3,4,5,6,7,8]$;
    \item AP$_{i}$: $\lambda_{i}=[0.1,0.15,0.2,0.25,0.3,\dots,0.5,0.55]$.
\end{itemize}
For the EP, only one parameter, called desired levels ($dl$), needs to be given, due to its automatic process. To extract the EPs in our case, we fully follow the same setup steps suggested in \cite{ghamisi2016extinction}; the $dl$ is assigned to be 7. More details can be found in \cite{ghamisi2016extinction} and \cite{rasti2017hyperspectral}.

Furthermore, the parameters in our IAPs consist of the number of scaled convolution kernels ($n_{s}$) and their radii of convolution kernels in spatially isotropic filtering ($r$), the number of Fourier orders ($m$), and the dimension in FL or DR ($d$) as well as the number of the grouped HSIs ($n_{g}$). In practical applications, these parameters can be determined by cross-validation on the available training set in order to achieve an automatic system for HSI feature extraction and classification. More specifically, the parameter combination is set to $n_s=3$, $r=[2,4,6]$, $m=[0,1,2,3]$, $d=30$, and $n_g=5$ for the Pavia University dataset, $n_s=3$, $r=[2,4,6]$, $m=[0,1,2,3]$, $d=30$, and $n_g=4$ for the Houston2013 dataset, and $n_s=3$, $r=[2,4,6]$, $m=[0,1,2]$, $d=40$, and $n_g=4$ for the Houston2018 dataset. Note that the dimension of the original IAPs without FL or DR is 341 for the first dataset, 396 for the second dataset, and 208 for the latter dataset in our case, which includes the original spectral signatures (103, 144, and 50), SIFs (103, 144, and 50), and FIFs (135, 108, and 84). In the following section, we experimentally discuss and analyze the various parameters used in the process of extracting the IAPs.
\begin{figure*}[!t]
	  \centering
		\subfigure{
			\includegraphics[width=0.8\textwidth]{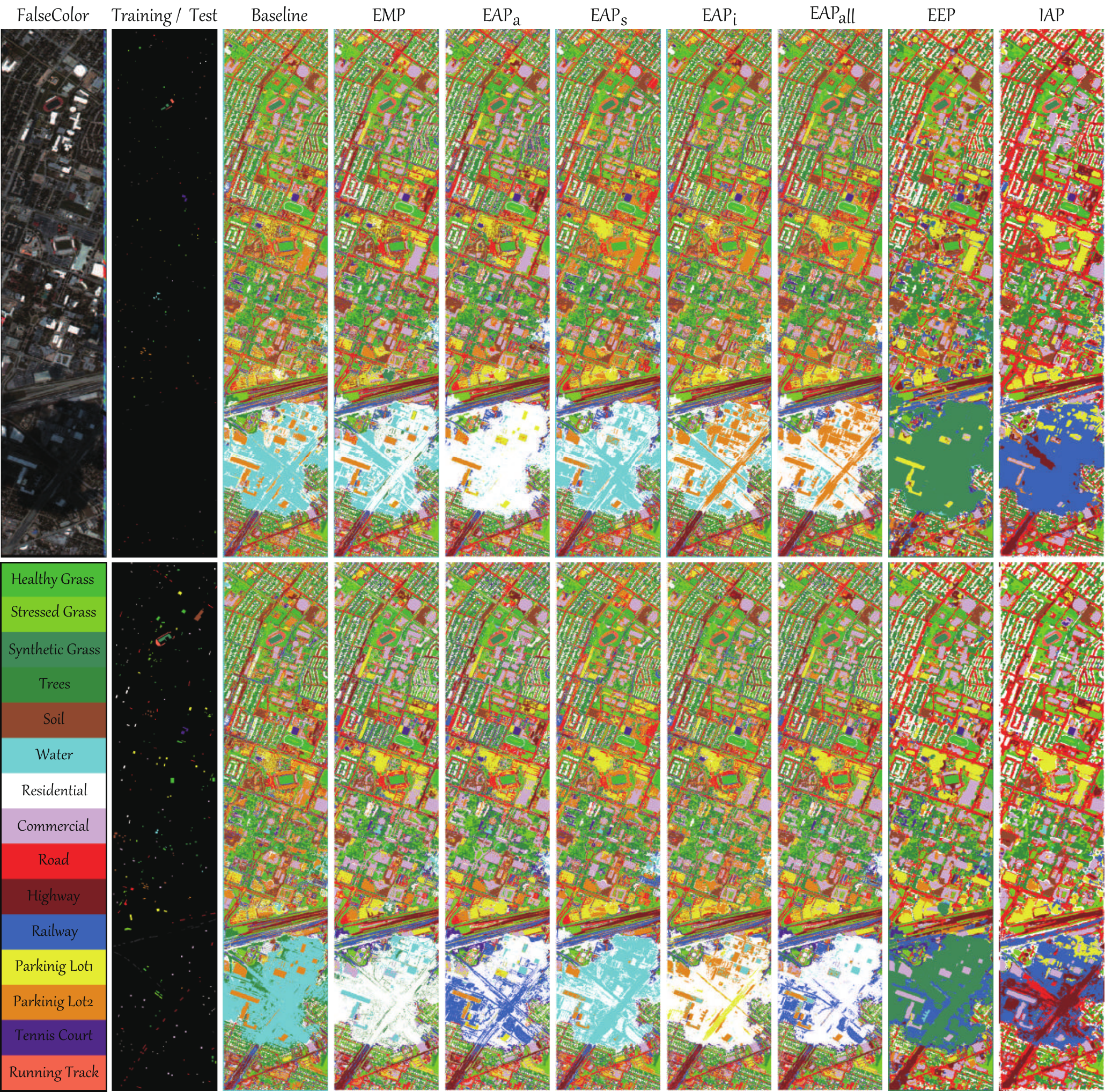}
		}
        \caption{Scene categories and visualization of false-color image, the distribution of training and test samples, and classification maps of different compared algorithms using two classifiers (top: NN, bottom: RF) on the Houston2013 dataset.}
\label{fig:CM_Houston2013}
\end{figure*}
\begin{table*}[!t]
\centering
\caption{Quantitative classification performance for eight different algorithms on the Houston2013 dataset}
\resizebox{\textwidth}{!}{
\begin{tabular}{c|cc|cc|cc|cc|cc|cc|cc|cc}
\toprule[1.5pt]Methods&\multicolumn{2}{c|}{Baseline (144) }&\multicolumn{2}{c|}{EMP (63) }&\multicolumn{2}{c|}{EAP$_a$ (63) }&\multicolumn{2}{c|}{EAP$_s$ (63) }&\multicolumn{2}{c|}{EAP$_i$ (63) }&\multicolumn{2}{c|}{EAP$_{all}$ (189) }&\multicolumn{2}{c|}{EEP (213)}&\multicolumn{2}{c}{IAP (30) }\\
\hline Classifier& NN & RF & NN & RF & NN & RF & NN & RF & NN & RF & NN & RF & NN & RF & NN & RF\\
\hline OA&72.83&73.11&77.45&77.90&77.79&78.17&76.91&77.78&75.25&77.08&79.34&79.01&78.74&81.80&83.86&\bf88.79\\
              AA&76.16&77.01&80.80&81.07&81.62&81.47&79.91&80.80&79.09&80.30&82.36&82.43&82.17&84.87&86.01&\bf90.09\\
              $\kappa$&0.7079&0.7110&0.7565&0.7606&0.7591&0.7633&0.7518&0.7608&0.7336&0.7519&0.7769&0.7724&0.7704&0.8035&0.8248&\bf0.8783\\
\hline Class1&82.15&82.62&82.34&81.67&80.34&82.05&81.77&82.05&80.06&82.05&81.86&82.05&80.63&81.67&81.29&\bf82.81\\
              Class2&81.86&83.36&82.52&82.80&83.27&82.61&83.93&80.55&83.83&78.48&84.12&80.45&83.93&84.02&83.27&\bf85.06\\
              Class3&99.60&97.82&99.01&99.60&\bf100.00&\bf100.00&\bf100.00&\bf100.00&\bf100.00&\bf100.00&\bf100.00&\bf100.00&99.21&99.80&\bf100.00&99.60\\
              Class4&\bf91.76&91.67&88.64&83.90&77.27&85.80&88.64&89.11&81.44&90.91&82.10&90.53&86.74&91.29&90.53&91.38\\
              Class5&97.06&96.50&99.62&98.48&99.24&98.39&97.73&96.59&96.88&94.60&99.34&95.93&99.62&97.63&99.62&\bf100.00\\
              Class6&95.10&\bf99.30&95.10&95.10&\bf99.30&99.30&95.10&94.41&96.50&95.10&92.31&98.60&95.10&94.41&97.20&98.60\\
              Class7&73.60&74.53&86.75&81.90&81.34&83.86&80.41&85.07&72.57&88.06&83.86&86.47&88.34&84.14&81.72&\bf85.35\\
              Class8&36.37&33.43&48.72&40.84&47.58&44.35&37.70&42.07&36.47&40.17&37.70&45.39&39.98&65.91&64.58&\bf74.45\\
              Class9&66.19&68.84&78.28&81.96&80.83&76.39&74.03&79.04&75.17&73.94&79.51&76.02&75.64&83.10&82.91&\bf83.76\\
              Class10&49.23&44.11&53.38&47.97&48.07&47.39&48.07&48.17&65.73&61.68&64.48&54.05&53.76&67.28&58.59&\bf94.02\\
              Class11&67.74&69.83&76.09&77.13&73.72&93.26&74.95&79.98&71.63&76.57&76.19&83.49&69.92&69.26&\bf94.02&90.32\\
              Class12&54.27&56.29&58.89&69.26&77.23&61.48&76.66&69.93&58.79&58.98&79.25&69.84&81.84&70.70&84.53&\bf90.39\\
              Class13&51.93&60.00&65.61&75.79&76.14&68.07&61.75&67.37&67.02&71.23&74.74&76.14&82.46&\bf89.12&73.33&76.14\\
              Class14&97.57&99.19&\bf100.00&\bf100.00&\bf100.00&\bf100.00&98.79&99.19&\bf100.00&97.98&\bf100.00&\bf100.00&95.95&99.60&99.60&99.60\\
              Class15&97.89&97.67&97.04&99.58&\bf100.00&99.15&99.15&98.52&\bf100.00&94.71&\bf100.00&97.46&99.37&95.14&98.94&99.79\\
\bottomrule[1.5pt]
\end{tabular}}
\label{tab:Houston2013}\\
\footnotesize{$^{*}$ The best is shown in bold. The feature dimensions are given in the parentheses after the method's name.}
\end{table*}
\begin{figure*}[!t]
	  \centering
		\subfigure{
			\includegraphics[width=0.72\textwidth]{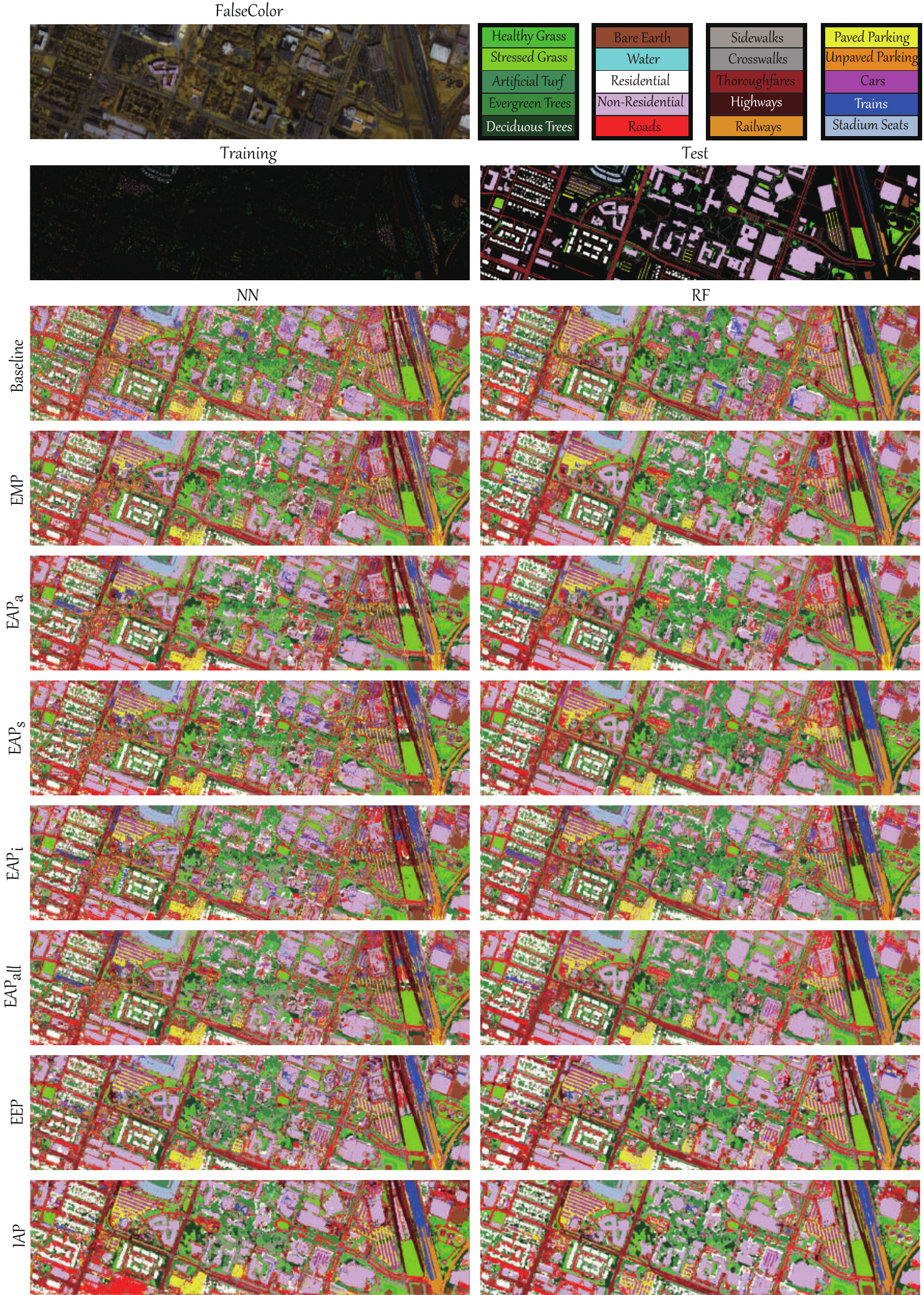}
		}
        \caption{Scene categories and visualization of false-color image, the distribution of training and test samples, and classification maps of different compared algorithms using two classifiers (NN and RF) on the Houston2018 dataset.}
\label{fig:CM_Houston2018}
\end{figure*}
\begin{table*}[!t]
\centering
\caption{Quantitative classification performance for eight different algorithms on the Houston2018 dataset.}
\resizebox{\textwidth}{!}{
\begin{tabular}{c|cc|cc|cc|cc|cc|cc|cc|cc}
\toprule[1.5pt]Methods&\multicolumn{2}{c|}{Baseline (50) }&\multicolumn{2}{c|}{EMP (63) }&\multicolumn{2}{c|}{EAP$_a$ (63) }&\multicolumn{2}{c|}{EAP$_s$ (63) }&\multicolumn{2}{c|}{EAP$_i$ (63) }&\multicolumn{2}{c|}{EAP$_{all}$ (189) }&\multicolumn{2}{c|}{EEP (213)}&\multicolumn{2}{c}{IAP (40) }\\
\hline Classifier& NN & RF & NN & RF & NN & RF & NN & RF & NN & RF & NN & RF & NN & RF & NN & RF\\
\hline OA&60.99&65.85&63.87&71.88&65.98&73.48&61.41&73.67&61.42&68.66&65.33&77.36&70.69&75.09&79.11&\bf83.68\\
              AA&77.07&81.09&81.16&85.69&82.21&85.99&80.10&85.55&78.58&82.64&82.38&87.78&86.63&88.55&\bf90.27&88.84\\
              $\kappa$&0.5358&0.5922&0.5689&0.6570&0.5912&0.6742&0.5421&0.6751&0.5426&0.6209&0.5851&0.7187&0.6471&0.6958&0.7408&\bf0.7931\\
\hline Class1&93.15&\bf95.03&91.95&93.29&91.69&94.16&91.49&93.72&90.84&93.15&91.26&93.99&91.33&93.66&90.38&93.73\\
              Class2&87.48&\bf91.35&87.71&90.39&87.60&90.02&84.46&89.21&83.48&90.46&86.60&90.98&89.96&90.70&86.27&89.20\\
              Class3&\bf100.00&\bf100.00&98.54&\bf100.00&\bf100.00&\bf100.00&\bf100.00&\bf100.00&96.10&\bf100.00&\bf100.00&\bf100.00&99.19&\bf100.00&\bf100.00&99.84\\
              Class4&91.31&94.64&93.92&94.84&92.85&95.17&90.46&93.64&92.28&94.28&93.96&95.31&95.07&95.31&97.65&\bf98.39\\
              Class5&87.36&90.22&90.11&92.52&94.33&94.85&91.82&93.54&91.16&92.85&93.65&95.05&96.26&95.40&\bf97.98&96.17\\
              Class6&98.28&98.35&97.81&98.97&99.68&99.93&99.58&99.80&99.19&99.85&99.80&\bf100.00&99.02&99.68&99.90&99.98\\
              Class7&92.92&90.83&98.33&98.33&98.75&98.75&95.42&98.33&90.83&96.67&98.33&98.33&97.92&97.92&\bf99.58&98.33\\
              Class8&67.03&82.24&83.52&89.78&86.12&89.72&82.63&89.64&80.55&86.42&85.93&90.80&85.74&\bf91.75&90.69&90.17\\
              Class9&58.46&60.59&57.63&66.79&61.30&70.42&55.22&71.72&54.96&64.14&59.39&75.02&62.97&68.69&77.01&\bf85.57\\
              Class10&41.30&46.42&44.61&55.44&48.32&58.41&44.36&57.23&45.78&54.57&47.72&62.59&59.61&62.81&65.48&\bf65.51\\
              Class11&33.74&44.47&41.23&50.07&44.25&52.36&37.40&45.93&43.29&46.07&47.03&54.73&49.00&56.32&55.52&\bf62.59\\
              Class12&37.51&31.79&45.13&47.84&56.41&51.36&50.62&46.23&45.35&37.58&52.53&50.40&67.77&62.42&\bf81.39&48.79\\
              Class13&43.89&50.80&51.48&63.28&49.43&59.19&44.13&61.10&45.60&52.21&49.47&68.91&68.22&71.68&75.04&\bf76.84\\
              Class14&84.32&86.20&86.62&91.77&82.98&91.40&85.36&92.41&80.30&86.20&86.46&95.46&96.66&95.87&\bf98.49&97.85\\
              Class15&98.12&98.07&98.66&98.90&96.36&98.18&98.79&99.75&93.24&98.14&96.16&\bf99.78&99.60&99.63&99.52&99.33\\
              Class16&86.82&89.09&86.62&93.43&84.78&90.21&86.76&92.65&84.56&92.00&86.76&93.94&95.21&97.51&\bf97.87&95.24\\
              Class17&95.56&91.85&\bf100.00&\bf100.00&\bf100.00&\bf100.00&\bf100.00&99.26&99.26&\bf100.00&\bf100.00&\bf100.00&98.52&\bf100.00&99.26&83.70\\
              Class18&69.69&87.87&79.85&89.98&79.25&88.10&77.41&88.98&75.60&82.79&81.29&90.61&86.9&93.73&94.92&\bf96.08\\
              Class19&80.10&94.12&91.12&99.03&90.61&97.97&86.87&98.34&81.21&86.70&91.68&\bf99.88&95.38&98.71&99.22&99.61\\
              Class20&94.43&97.93&98.39&99.21&99.49&99.67&99.16&99.60&98.09&98.80&99.54&99.86&98.24&99.24&99.30&\bf99.95\\
\bottomrule[1.5pt]
\end{tabular}}
\label{tab:Houston2018}\\
\footnotesize{$^{*}$ The best is shown in bold. The feature dimensions are given in the parentheses after the method's name.}
\end{table*}

\subsection{Results and Discussion}
\subsubsection{Pavia University Dataset}The classification maps and corresponding quantitative results in terms of \textit{OA}, \textit{AA}, and \textit{$\kappa$} for different compared methods using NN and RF classifiers are given in Fig. \ref{fig:CM_Pavia} and Table \ref{tab:pavia}, respectively.

In detail, the classification accuracies using extracted features are much higher than that of using only OSF (at least 7\% increase) for both classifiers. For example, the EMPs can effectively represent the structure information of objects in the HSI, yielding competitive classification performance even better than many single-attribute EAPs, such as EAPs$_s$, EAPs$_i$. By focusing on HSI's spatial information, the EAP$_a$ method is able to excavate the regional features, leading to a more smooth classification result (see Fig. \ref{fig:CM_Pavia}). As expected, collecting multiple attributes into a stacked vector representation (EAPs$_{all}$) achieves a great improvement in classification performance (around 5\% in terms of OAs) compared to the single-attribute one. However, the use of extinction filters makes another breakthrough on the basis of AF-based profiles, with an additional 2\% increase in terms of OAs using the two classifiers (EEPs \textit{versus} EAPs$_{all}$).

Although the above-mentioned methods have shown superiority in spatial information modeling and further obtained excellent performance on the pixel-wise classification task, yet their ability in addressing various spectral variabilities and semantic changes of local scene or objects is relatively weak. In contrast, the proposed IAP method is capable of extracting the intrinsic invariant feature representation from the HSI, achieving a more effective feature extraction. As shown in Fig. \ref{fig:CM_Pavia}, there are less noisy points in the classification maps obtained by our method. In particular, the \textit{Meadows} class located in the bottom of the studied scene shows a more reasonable result, which is approaching to the manually-labeled ground truth. Similarly, the IAP approach dramatically outperforms the others with a quantitative comparison in Table \ref{tab:pavia}.

\subsubsection{Houston2013 Dataset} Fig. \ref{fig:CM_Houston2013} visualizes the classification maps of different feature extraction methods using two classifiers on the Houston2013 dataset, while Table \ref{tab:Houston2013} quantifies the corresponding experimental results in terms of \textit{OA}, \textit{AA}, and \textit{$\kappa$} as well as the classification accuracy for each class.

Generally speaking, the classification performance with the feature extraction step is obviously superior to that with the OSF (Baseline), which is consistent for both NN and RF classifiers. Although the EMPs yield a similar performance compared to the EAPs with a single attribute (e.g., EAPs$_a$, EAPs$_s$, and EAPs$_i$) using the two classifiers, it is worth noting that those single-attribute EAPs usually hold a lower feature dimension. This indicates, to some extent, that EAPs would have the advantages over EMPs in extracting the spatial information of the image, since EAPs-based methods could make it possible to model more geometrical features (e.g., size, shape, texture). As expected, stacking all single EAPs is of great benefit in finding a more discriminative feature representation; thus the resulting EAPs$_{all}$ performs better at classifying each material than any single EAP or EMP. With the use of extinction filters, the developed EEPs outperform the aforementioned methods. Particularly when using the RF classifier, there is an obvious improvement in classification accuracy, improving the OAs by approximately 2\% and 5\%, over the EAPs$_{all}$ and EMPs, respectively.

Remarkably, our proposed IAP method not only outperforms other feature extraction operators overall in terms of \textit{OA}, \textit{AA}, and $\kappa$, but also it achieves the desirable results for each class, especially for challenging classes like \textit{Commercial}, \textit{Highway}, and \textit{Parking Lot1}. For these classes, the IAPs make a significant performance improvement, at an increase of at least 10\% accuracy. We can also observe from Fig. \ref{fig:CM_Houston2013} that the classification maps obtained by the proposed feature extractor are more smooth in the regions sharing the same materials and sharper on the edges between different materials. Furthermore, there are some object deformations (e.g., shift, rotation) in the stadium and its surroundings located in the middle of the studied HSI. Owing to the robustness in local semantic changes of the scene, the IAPs obtain more accurate classification maps in the area in and around the stadium.
\begin{figure*}[!t]
	  \centering
	  	\subfigure[Pavia University Dataset]{
			\includegraphics[width=0.98\textwidth]{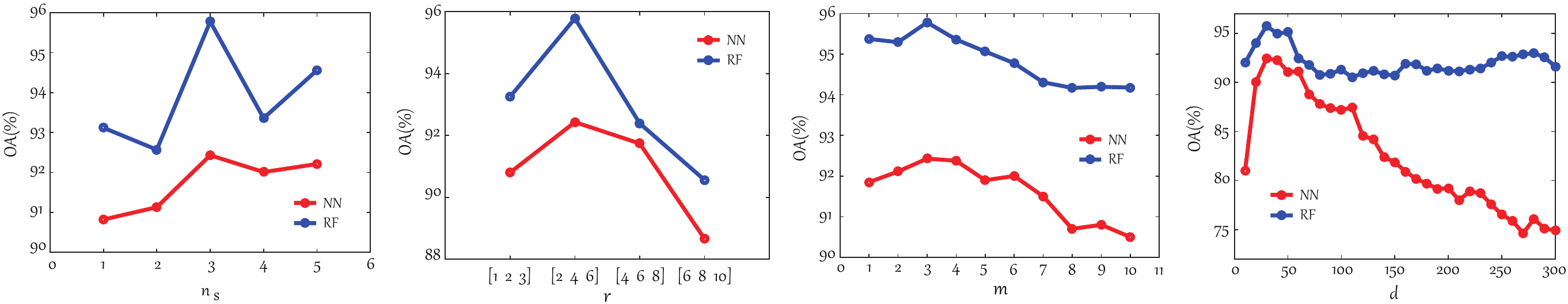}
		}
		\subfigure[Houston2013 Dataset]{
			\includegraphics[width=0.98\textwidth]{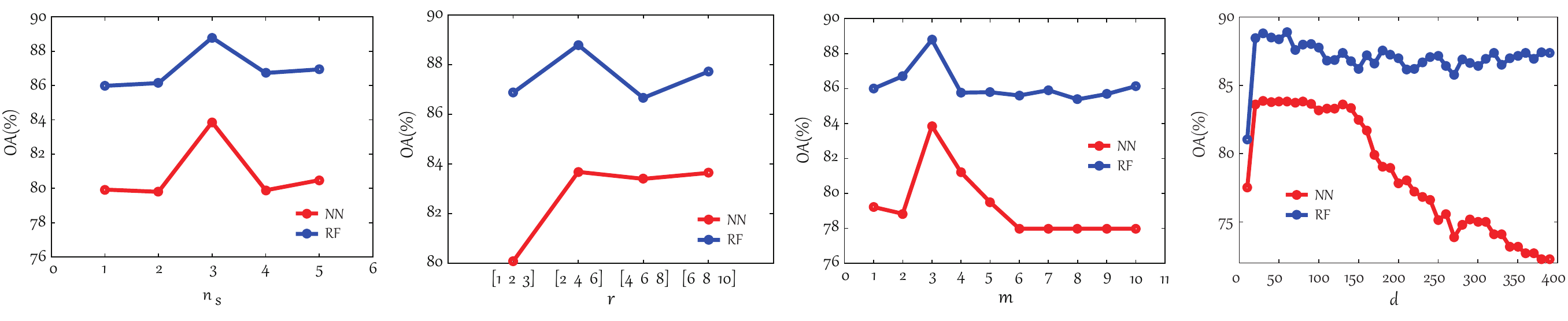}
		}
		\subfigure[Houston2018 Dataset]{
			\includegraphics[width=0.98\textwidth]{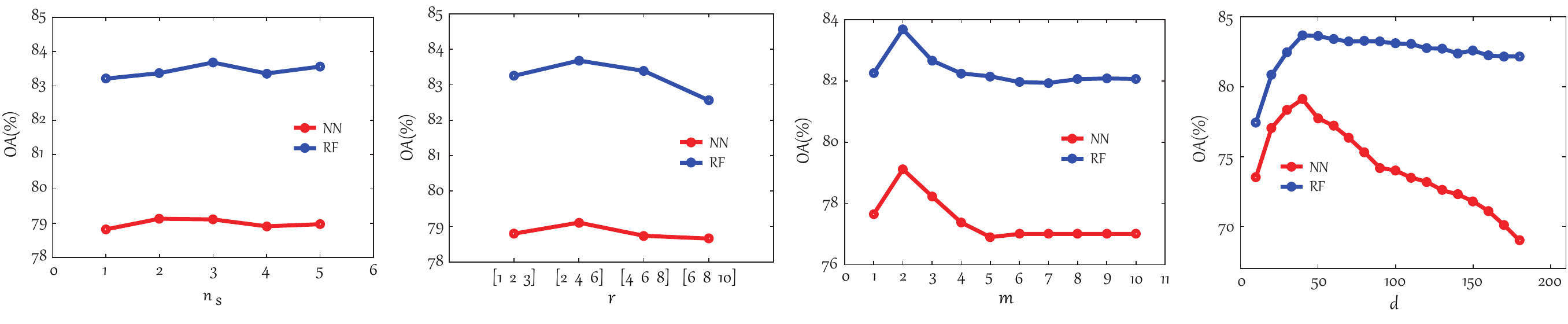}
		}
        \caption{Parameter sensitivity analysis of the proposed IAPs with respect to the number of scaled convolution kernels ($n_s$) and their radii of these convolution kernels ($r$) in spatially isotropic filtering, the number of Fourier orders (m), and feature dimension ($d$) after FL or DR, respectively. (a) Analysis results of two classifiers (NN and RF) in terms of OAs on the Pavia University dataset; (b) Analysis results of two classifiers (NN and RF) in terms of OAs on the Houston2013 dataset; (c) Analysis results of two classifiers in terms of OAs on the Houston2018 dataset.}
\label{fig:para}
\end{figure*}

\subsubsection{Houston2018 Dataset}
For the Houston2018 dataset, we make a similar visual comparison for eight compared approaches, as given in Fig. \ref{fig:CM_Houston2018}; the specific numerical results are detailed in Table \ref{tab:Houston2018}.

As observed in Table \ref{tab:Houston2018}, the same basic trend appears when using NN and RF for all candidates: that is, the classification accuracies using RF are higher than those using NN. More specifically, with the original spectral features as the input, the baseline only holds 60.99\% and 65.85\% results using NN and RF classifiers, respectively, due to the lack of spatial information modeling. In MPs, the spatial information is considered in the form of openings and closings, yet the EMPs yield a relatively poor classification performance. This might result from some intrinsic limitations of MPs. For example, the MPs are not capable of adaptively making an effective connection between different scaled objects, further limiting their ability to characterize the semantic structural information. Despite the use of mathematical morphological attribute filers that are able to generate a richer geometrical description, EAP-related approaches fail to improve classification performance dramatically, since there are more challenging categories and more complex semantic variations in the studied scene. Nevertheless, the ability of EAPs$_{all}$ to fuse the different attributes still contributes to its power in spatial information extraction, yielding an increase in accuracy of around 5\% using the RF classifier, compared to the others previously mentioned. Unlike EMPs and EAPs, EEPs utilize a sequence of extrema-oriented connected filters to extract the extinction values as the features. Due to these constructed features, the classification accuracy of EEPs increases by approximately 5\% using NN, in spite of a slight decrease in using RF (about 2 percentage points less than EAPs$_{all}$).

Beyond those algorithms developed based on MPs and APs, we propose to model the invariant features from both the spatial and frequency domains, which has been theoretically proven to be robust against the semantic change caused by various image deformations (e.g., shift, rotation, sensor noises, or distortions). Therefore, the resulting IAPs outperform the previously-proposed MPs or APs methods, showing the best overall performance and the highly competitive classification results for each class, as shown in Table \ref{tab:Houston2018}.

Fig. \ref{fig:CM_Houston2018} also highlights the superiority of the proposed IAP method by means of classification maps. Generally speaking, our method tends to lead to more smooth classification maps: that is, the IAPs aggregate the same materials more easily while separating the different materials. Our IAPs also remove the effects of hot pixels like salt-and-pepper noise from classification maps effectively and simultaneously preserve the semantically meaningful structure or objects, leading to an increase of about 20\% in classification accuracy. This is particularly the case for the class of Non-Residential Buildings that constitute large-scale coverage of the whole scene.

\begin{table*}[!t]
\centering
\caption{Classification performance in terms of OA for different studied methods with DR techniques on the three datasets. Note that the optimal dimension for each method is selected by the means of a cross-validation strategy.}
\resizebox{\textwidth}{!}{
\begin{tabular}{c|cc|cc|cc|cc|cc|cc|cc|cc}
\toprule[1.5pt]
Methods&\multicolumn{2}{c|}{Baseline }&\multicolumn{2}{c|}{EMP}&\multicolumn{2}{c|}{EAP$_a$ }&\multicolumn{2}{c|}{EAP$_s$}&\multicolumn{2}{c|}{EAP$_i$}&\multicolumn{2}{c|}{EAP$_{all}$}&\multicolumn{2}{c|}{EEP}&\multicolumn{2}{c}{IAP}\\
\hline Classifier& NN & RF & NN & RF & NN & RF & NN & RF & NN & RF & NN & RF & NN & RF & NN & RF\\
\hline Dimension & \multicolumn{2}{c|}{(30,\; 20,\; 30)} & \multicolumn{2}{c|}{(40,\; 30,\; 30)} & \multicolumn{2}{c|}{(30,\; 20,\; 20)} & \multicolumn{2}{c|}{(20,\; 20,\; 30)} & \multicolumn{2}{c|}{(40,\; 30,\; 20)} & \multicolumn{2}{c|}{(20,\; 40,\; 30)} & \multicolumn{2}{c|}{(30,\; 30,\; 20)} & \multicolumn{2}{c}{(30,\; 30,\; 40)}\\
\hline
            Pavia Univeristy&71.75&77.50&78.04&80.69&79.75&82.23&78.50&84.47&67.15&74.26&82.67&87.35&85.64&91.06&92.43&\bf95.78\\
            Houston2013&72.78&83.75&77.31&80.31&78.00&81.32&76.99&77.91&74.87&77.07&79.23&80.06&78.56&81.50 &83.86&\bf88.79\\
            Houston2018&60.98&72.75&63.83&66.47&65.27&69.95&63.12&72.35&60.99&65.33&65.16&73.39&70.62&69.46& 79.11&\bf83.68\\
\bottomrule[1.5pt]
\end{tabular}
}
\label{tab:classificaiton_DR}
\end{table*}

\begin{table*}[!t]
\centering
\caption{Ablation analysis of the proposed IAPs in terms of OA on the three datasets. Dimension with different component combination is also shown.}
\resizebox{0.8\textwidth}{!}{
\begin{tabular}{cccccccccccc}
\toprule[1.5pt]
\multirow{2}{*}{Methods} & \multirow{2}{*}{OSF} & \multirow{2}{*}{SIFs} &\multirow{2}{*}{FIFs} &\multirow{2}{*}{FL or DR} & \multirow{2}{*}{Dimension} & \multicolumn{2}{c}{Pavia University} & \multicolumn{2}{c}{Houston2013} & \multicolumn{2}{c}{Houston2018} \\
\cline{7-12} & & & & & & NN & RF & NN & RF & NN & RF\\
\hline\hline
IAPs & $\checkmark$ & $\times$ & $\times$ & $\times$ & (103,\; 144,\; 50) & 71.85 & 71.50 & 72.83 & 73.11 & 60.99 & 65.85\\
IAPs & $\times$ & $\checkmark$ & $\times$ & $\times$ & (103,\; 144,\; 50) & 70.77 & 82.48 & 71.21 & 82.89 & 67.94 & 74.23\\
IAPs & $\times$ & $\times$ & $\checkmark$ & $\times$ & (135,\; 108,\; 84) & 70.46 & 76.37 & 40.44 & 63.79 & 35.57 & 52.27\\
IAPs & $\checkmark$ & $\checkmark$ & $\times$ & $\times$ & (206,\; 288,\; 100) & 72.48 & 86.63 & 72.35 & 83.67 & 64.95 & 79.29\\
IAPs & $\checkmark$ & $\checkmark$ & $\checkmark$ & $\times$ & (341,\; 396,\; 184) & 81.16 & 90.33 & 74.46& 87.05 & 69.04 & 81.74\\
IAPs & $\checkmark$ & $\checkmark$ & $\checkmark$ & $\checkmark$ & (30,\; 30,\; 40) & 92.43 & 95.78 & 83.86 & 88.79 & 79.11 & 83.68\\
\bottomrule[1.5pt]
\end{tabular}
}
\label{table:Ablation}\\
\footnotesize{$\checkmark$ and $\times$ mean the current component with and without being involved or considered in the HSI classification framework, respectively.}
\end{table*}

From the experimental results on the different datasets, we can observe an interesting and meaningful phenomenon that the proposed IAPs are more apt to recognize objects or materials with a regular structure (or shape), such as Road, Residential, and Commercial, while for those irregular classes (e.g., grass and tree) that seem to hold more disorderly textual features, the extracted IAPs might not be so discriminative to identify these materials, limiting the gain in the final classification performance to some extent.

\subsection{Parameter Sensitivity Analysis}
The effectiveness of a feature extractor largely depends on parameter selection. It is therefore indispensable to discuss the sensitivity of parameters involved in the proposed IAP. These parameters include the number of scaled convolution kernels ($n_s$) and the corresponding radii of these convolution kernels ($r$) in spatially isotropic filtering, the number of Fourier orders ($m$) in the process of extracting frequency features, and the reduced dimension ($d$) in FL or DR. Among them,
\begin{itemize}
    \item the search range of $n_s$ is constrained from 1 to 5;
    \item the $r$ can be investigated through the combination of $[1,2,3]$, $[2,4,6]$, $[4,6,8]$, and $[6,8,10]$ \footnote{In the original literature \cite{liu2014rotation}, $n_s=3$ is suggested. Also, we experimentally found in our case that the $n_s$ is equal to 3, yielding the best performance on both datasets.};
    \item the order $m$ can be selected from the different set, i.e., $[0, 1]$, $[0,1,2]$, $[0,1,2,3]$, $[0,1,2,3,4]$, $[0,1,2,3,4,5]$, $\dots$, $[0,1,2,\dots,9,10]$;
    \item the feature dimension after FL or DR is determined ranging from 10 to the dimension of the original IAPs (e.g., 300 for the Pavia University, 390 for the Houston2013 and 180 for the Houston2018) at a 10 interval.
\end{itemize}

We experimentally analyze the effects of different parameters for the overall classification accuracy (OA in our case) with two classifiers. The resulting quantitative results are shown in Fig. \ref{fig:para} for the three datasets.

It is evident from Fig. \ref{fig:para} that the optimal parameter combinations are $n_s=3$, $r=[2,4,6]$, $m=[0,1,2,3]$, and $d=30$  for the Pavia University dataset, $n_s=3$, $r=[2,4,6]$, $m=[0,1,2,3]$, and $d=30$ for the Houston2013 dataset, and $n_s=3$, $r=[2,4,6]$, $m=[0,1,2]$, and $d=40$ for the Houston2018. Moreover, we also discovered an interesting phenomenon: classification performance is insensitive to the parameters of $n_s$ and $r$ on the Houston2018 datasets, while for the Pavia University and Houston2013 they have a moderate effect on OAs. Notably, the Fourier orders ($m$) are associated with classification results, but a progressively increased $m$ starts with a sharp performance decrease and then becomes stable with a relatively poor result. This reveals that $m$ is a noteworthy and important parameter during the process of extracting IAPs. The graph showing results for the parameter $d$ in Fig. \ref{fig:para} demonstrates that the extracted IAPs are, to some extent, redundant and can be further improved in feature discriminative ability by means of FL or DR, as the OAs start to quickly reach a certain optimal value (e.g., 30 for the Pavia University dataset, 30 for the Houston2013 dataset, and 40 for the Houston2018 dataset), then hold basically unchanged over a period of the time, and finally gradually diminish for the robust RF classifier. For the NN classifier, the OAs dramatically decrease when inputting the IAPs with a higher dimension. This could be result of the curse of dimensionality.

\noindent \textbf{Effects of DR or FL in Feature Extraction:} For a fair comparison, we also investigate the effects of a DR technique on all studied methods, as listed in Table \ref{tab:classificaiton_DR}.

There is an interesting regular pattern in Table \ref{tab:classificaiton_DR}. Generally, the classification performance using the dimension-reduced features is obviously superior to that using original features (without DR), e.g., Baseline and our IAP method. It should be noted that, however, those EsP-based approaches with a DR technique usually perform worse than their previous versions without DR. A possible reason is the use of a DR technique (e.g., PCA) before feature extraction. If the DR technique is applied again after feature extraction, then this might suffer from the effects of reuse, leading to the limitation and even degradation in the final classification performance.

\subsection{Ablation Studies}
In addition, we investigate and analyze the performance improvement of our IAP method by step-wise adding the different components, since the IAPs are involved with multiple components, i.e., SIFs, FIFs, and FL or DR. Table VIII shows an increasing performance gain as the different techniques are gradually fused.  We found that successively adding each component into the proposed IAPs led to a progressive enhancement in feature representation ability, yielding a higher classification result for the HSI. This also indicates the reasonableness and advancement of the proposed IAPs-based HSI classification framework we designed (see the Section II.D and Fig. \ref{fig:Classification}).

The key points from the ablation analysis can be summarized as follows:
\begin{itemize}
    \item[1)] There is a great improvement in classification performance after applying the FL or DR on the original IAPs, from which we conclude that the FL or DR step plays a significant role in our proposed IAPs-based HSI classification framework. This might be explained by the fact that directly stacking the OSF, SIFs, and FIFs as a final input would hurt classification performance, since they come from different feature spaces. Moreover, this is also due to the reason that FL or DR can address the curse of dimensionality (the lack of balance between the number of features and training samples) as well as the existence of high redundancy among IAPs.
    \item[2)] Although the results with SIFs are superior to those with FIFs alone, performance is still limited without FIFs, showing the power of FIFs in enriching the diversity and robustness of the extracted features, particularly when they are jointly used, i.e., IAPs (OSF+SIFs+FIFs).
    \item[3)] Using FIFs individually would yield poor results, which indicates that frequency information alone is not sufficiently discriminative to identify a variety of materials.
\end{itemize}

\subsection{Computational Cost in the Proposed IAP}
All experiments in this paper were implemented with Matlab 2016a on a Windows 10 operation system and conducted on Intel Core i7-6700HQ 2.60GHZ laptop with 16GB memory. With the setting, the running time of the proposed IAP is given in Table VIII. Overall, the running time is acceptable in practice and it shows an approximately linear increase as the size of the image increases.
\begin{table}[!h]
\centering
\caption{Processing time for the proposed IAP on the three datasets}
\resizebox{0.46\textwidth}{!}{
\begin{tabular}{c|c|c|c}
\toprule[1.5pt] Datasets & Pavia University & Houston2013 & Houston2018\\
\hline Running Time (s) & 11.2 & 29.0 & 62.7\\
\bottomrule[1.5pt]
\end{tabular}
}
\label{tab:timecost}
\end{table}

\section{Conclusion}
To effectively improve the robustness of a feature extractor in modeling spatial information of HSI, we propose a novel AP-like feature descriptor called IAP that designs the invariant APs in both spatial and frequency domains. The proposed IAP method aims at overcoming the shortcomings of previous MPs or APs, in which extracted spatial features are apt to be degraded, leading to a large difference between the same materials, or confusion with other different materials, due to the local semantic change in a hyperspectral scene. Combined with the sound theory in modeling our proposed SIFs and FIFs, the resulting IAPs have demonstrated their potential and superiority in the HSI classification task. In the future, we will further extend our model to a supervised or semi-supervised version in an end-to-end fashion (e.g., deep learning) to conceive invariant feature extraction and classification as a whole.
\bibliographystyle{IEEEbib}
\bibliography{HDF_ref}

\begin{thebibliography}{10}

\bibitem{anderson1976land}
J.~R. Anderson,
\newblock {\em A land use and land cover classification system for use with
  remote sensor data}, vol. 964,
\newblock US Government Printing Office, 1976.

\bibitem{kang2018building}
J.~Kang, M.~K{\"o}rner, Y.~Wang, H.~Taubenb{\"o}ck, and X.~X. Zhu,
\newblock ``Building instance classification using street view images,''
\newblock {\em ISPRS J. Photogramm. Remote Sens.}, vol. 145, pp. 44--59, 2018.

\bibitem{li2017hyperspectral}
H.~Li, Y.~Song, and P.~Chen,
\newblock ``Hyperspectral image classification based on multiscale spatial
  information fusion,''
\newblock {\em IEEE Trans. Geosci. Remote Sens.}, vol. 55, no. 9, pp.
  5302--5312, 2017.

\bibitem{hang2019cascaded}
R.~Hang, Q.~Liu, D.~Hong, and P.~Ghamisi,
\newblock ``Cascaded recurrent neural networks for hyperspectral image
  classification,''
\newblock {\em IEEE Trans. Geosci. Remote Sens.}, vol. 57, no. 8, pp.
  5384--5394, 2019.

\bibitem{gao2019spectral}
Q.~Gao, S.~Lim, and X.~Jia,
\newblock ``Spectral-spatial hyperspectral image classification using a
  multiscale conservative smoothing scheme and adaptive sparse
  representation,''
\newblock {\em IEEE Trans. Geosci. Remote Sens.}, vol. 57, no. 10, pp.
  7718--7730, 2019.

\bibitem{yedemir2019supervised}
Z.~Ye, J.~Chen, H.~Li, Y.~Wei, G.~Xiao, and J.~Benediktsson,
\newblock ``Supervised functional data discriminant analysis for hyperspectral
  image classification,''
\newblock {\em IEEE Trans. Geosci. Remote Sens.}, 2019,
\newblock DOI:10.1109/TGRS.2019.2940991.

\bibitem{hang2015matrix}
R.~Hang, Q.~Liu, H.~Song, and Y.~Sun,
\newblock ``Matrix-based discriminant subspace ensemble for hyperspectral image
  spatial--spectral feature fusion,''
\newblock {\em IEEE Trans. Geosci. Remote Sens.}, vol. 54, no. 2, pp. 783--794,
  2015.

\bibitem{hong2019cospace}
D.~Hong, N.~Yokoya, J.~Chanussot, and X.~X. Zhu,
\newblock ``Co{S}pace: Common subspace learning from
  hyperspectral-multispectral correspondences,''
\newblock {\em IEEE Trans. Geosci. Remote Sens.}, vol. 57, no. 7, pp.
  4349--4359, 2019.

\bibitem{hu2019comparative}
J.~Hu, D.~Hong, Y.~Wang, and X.~Zhu,
\newblock ``A comparative review of manifold learning techniques for
  hyperspectral and polarimetric sar image fusion,''
\newblock {\em Remote Sens.}, vol. 11, no. 6, pp. 681, 2019.

\bibitem{hong2019learnable}
D.~Hong, N.~Yokoya, N.~Ge, J.~Chanussot, and X.~X. Zhu,
\newblock ``Learnable manifold alignment ({L}e{MA}): A semi-supervised
  cross-modality learning framework for land cover and land use
  classification,''
\newblock {\em ISPRS J. Photogramm. Remote Sens.}, vol. 147, pp. 193--205,
  2019.

\bibitem{li2018real}
C.~Li, L.~Gao, Y.~Wu, B.~Zhang, J.~Plaza, and A.~Plaza,
\newblock ``A real-time unsupervised background extraction-based target
  detection method for hyperspectral imagery,''
\newblock {\em J. Real-Time Image Pr.}, vol. 15, no. 3, pp. 597--615, 2018.

\bibitem{xu2018joint}
Y.~Xu, Z.~Wu, J.~Chanussot, and Z.~Wei,
\newblock ``Joint reconstruction and anomaly detection from compressive
  hyperspectral images using mahalanobis distance-regularized tensor rpca,''
\newblock {\em IEEE Trans. Geosci. Remote Sens.}, vol. 56, no. 5, pp.
  2919--2930, 2018.

\bibitem{wu2019approximate}
Y.~Wu, S.~L{\'o}pez, B.~Zhang, F.~Qiao, and L.~Gao,
\newblock ``Approximate computing for onboard anomaly detection from
  hyperspectral images,''
\newblock {\em J. Real-Time Image Pr.}, vol. 16, no. 1, pp. 99--114, 2019.

\bibitem{hong2017unmixing}
D.~Hong, N.~Yokoya, J.~Chanussot, and X.~X. Zhu,
\newblock ``Learning a low-coherence dictionary to address spectral variability
  for hyperspectral unmixing,''
\newblock in {\em Proc. ICIP}. IEEE, Sep. 2017, pp. 1--5.

\bibitem{hong2018sulora}
D.~Hong and X.~X. Zhu,
\newblock ``S{UL}o{RA}: Subspace unmixing with low-rank attribute embedding for
  hyperspectral data analysis,''
\newblock {\em IEEE J. Sel. Topics Signal Process.}, vol. 12, no. 6, pp.
  1351--1363, 2018.

\bibitem{tang2018multiharmonic}
M.~Tang, B.~Zhang, A.~Marinoni, L.~Gao, and P.~Gamba,
\newblock ``Multiharmonic postnonlinear mixing model for hyperspectral
  nonlinear unmixing,''
\newblock {\em IEEE Geosci. Remote Sens. Lett.}, vol. 15, no. 11, pp.
  1765--1769, 2018.

\bibitem{hong2019augmented}
D.~Hong, N.~Yokoya, J.~Chanussot, and X.~X. Zhu,
\newblock ``An augmented linear mixing model to address spectral variability
  for hyperspectral unmixing,''
\newblock {\em IEEE Trans. Image Process.}, vol. 28, no. 4, pp. 1923--1938,
  2019.

\bibitem{liao2017taking}
W.~Liao, J.~Chanussot, M.~Dalla Mura, X.~Huang, R.~Bellens, S.~Gautama, and
  W.~Philips,
\newblock ``Taking optimal advantage of fine spatial resolution: Promoting
  partial image reconstruction for the morphological analysis of
  very-high-resolution images,''
\newblock {\em IEEE Geosci. Remote Sens. Mag.}, vol. 5, no. 2, pp. 8--28, 2017.

\bibitem{soille2003morphological}
P.~Soille,
\newblock ``Morphological image analysis: Principles and applications,''
\newblock 2003.

\bibitem{pesaresi2001new}
M.~Pesaresi and J.~A. Benediktsson,
\newblock ``A new approach for the morphological segmentation of
  high-resolution satellite imagery,''
\newblock {\em IEEE Trans. Geosci. Remote Sens.}, vol. 39, no. 2, pp. 309--320,
  2001.

\bibitem{benediktsson2005classification}
J.~A. Benediktsson, J.~A. Palmason, and J.~R. Sveinsson,
\newblock ``Classification of hyperspectral data from urban areas based on
  extended morphological profiles,''
\newblock {\em IEEE Trans. Geosci. Remote Sens.}, vol. 43, no. 3, pp. 480--491,
  2005.

\bibitem{wold1987principal}
S.~Wold, K.~Esbensen, and P.~Geladi,
\newblock ``Principal component analysis,''
\newblock {\em Chemom. Intell. Lab. Syst.}, vol. 2, no. 1-3, pp. 37--52, 1987.

\bibitem{fauvel2007spectral}
M.~Fauvel, J.~Chanussot, J.~A. Benediktsson, and J.~R. Sveinsson,
\newblock ``Spectral and spatial classification of hyperspectral data using
  svms and morphological profiles,''
\newblock in {\em Proc. IGARSS}. IEEE, 2007, pp. 4834--4837.

\bibitem{licciardi2012linear}
G.~Licciardi, P.~R. Marpu, J.~Chanussot, and J.~A. Benediktsson,
\newblock ``Linear versus nonlinear pca for the classification of hyperspectral
  data based on the extended morphological profiles,''
\newblock {\em IEEE Geosci. Remote Sens. Lett.}, vol. 9, no. 3, pp. 447--451,
  2012.

\bibitem{fauvel2013advances}
M.~Fauvel, Y.~Tarabalka, J.~A. Benediktsson, J.~Chanussot, and J.~C. Tilton,
\newblock ``Advances in spectral-spatial classification of hyperspectral
  images,''
\newblock {\em Proc. IEEE}, vol. 101, no. 3, pp. 652--675, 2013.

\bibitem{dalla2010morphological}
M.~Dalla Mura, J.~A. Benediktsson, b.~Waske, and L.~Bruzzone,
\newblock ``Morphological attribute profiles for the analysis of very high
  resolution images,''
\newblock {\em IEEE Trans. Geosci. Remote Sens.}, vol. 48, no. 10, pp.
  3747--3762, 2010.

\bibitem{breen1996attribute}
E.~J. Breen and R.~Jones,
\newblock ``Attribute openings, thinnings, and granulometries,''
\newblock {\em Comput. Vis. Image Und.}, vol. 64, no. 3, pp. 377--389, 1996.

\bibitem{hong2015novel}
D.~Hong, W.~Liu, J.~Su, Z.~Pan, and G.~Wang,
\newblock ``A novel hierarchical approach for multispectral palmprint
  recognition,''
\newblock {\em Neurocomputing}, vol. 151, pp. 511--521, 2015.

\bibitem{dalla2011classification}
M.~Dalla Mura, A.~Villa, J.~A. Benediktsson, J.~Chanussot, and L.~Bruzzone,
\newblock ``Classification of hyperspectral images by using extended
  morphological attribute profiles and independent component analysis,''
\newblock {\em IEEE Geosci. Remote Sens. Lett.}, vol. 8, no. 3, pp. 542--546,
  2011.

\bibitem{song2014remotely}
B.~Song, J.~Li, M.~Dalla Mura, P.~Li, A.~Plaza, J.~Bioucas-Dias, J.~A.
  Benediktsson, and J.~Chanussot,
\newblock ``Remotely sensed image classification using sparse representations
  of morphological attribute profiles,''
\newblock {\em IEEE Trans. Geosci. Remote Sens.}, vol. 52, no. 8, pp.
  5122--5136, 2014.

\bibitem{ghamisi2015survey}
P.~Ghamisi, M.~Dalla Mura, and J.~A. Benediktsson,
\newblock ``A survey on spectral--spatial classification techniques based on
  attribute profiles,''
\newblock {\em IEEE Trans. Geosci. Remote Sens.}, vol. 53, no. 5, pp.
  2335--2353, 2015.

\bibitem{xu2012morphological}
Y.~Xu, T.~G{\'e}raud, and L.~Najman,
\newblock ``Morphological filtering in shape spaces: Applications using
  tree-based image representations,''
\newblock in {\em Proc. ICPR}. IEEE, 2012, pp. 485--488.

\bibitem{falco2015spectral}
N.~Falco, J.~Benediktsson, and L.~Bruzzone,
\newblock ``Spectral and spatial classification of hyperspectral images based
  on ica and reduced morphological attribute profiles,''
\newblock {\em IEEE Trans. Geosci. Remote Sens.}, vol. 53, no. 11, pp.
  6223--6240, 2015.

\bibitem{demir2015histogram}
B.~Demir and L.~Bruzzone,
\newblock ``Histogram-based attribute profiles for classification of very high
  resolution remote sensing images,''
\newblock {\em IEEE Trans. Geosci. Remote Sens.}, vol. 54, no. 4, pp.
  2096--2107, 2015.

\bibitem{cavallaro2016remote}
G.~Cavallaro, M.~Dalla Mura, J.~Benediktsson, and A.~Plaza,
\newblock ``Remote sensing image classification using attribute filters defined
  over the tree of shapes,''
\newblock {\em IEEE Trans. Geosci. Remote Sens.}, vol. 54, no. 7, pp.
  3899--3911, 2016.

\bibitem{cavallaro2017automatic}
G.~Cavallaro, N.~Falco, M.~Dalla Mura, and J.~Benediktsson,
\newblock ``Automatic attribute profiles,''
\newblock {\em IEEE Trans. on Image Process.}, vol. 26, no. 4, pp. 1859--1872,
  2017.

\bibitem{liao2016morphological}
W.~Liao, M.~Dalla Mura, J.~Chanussot, R.~Bellens, and W.~Philips,
\newblock ``Morphological attribute profiles with partial reconstruction,''
\newblock {\em IEEE Trans. Geosci. Remote Sens.}, vol. 54, no. 3, pp.
  1738--1756, 2016.

\bibitem{ghamisi2016extinction}
P.~Ghamisi, R.~Souza, J.~A. Benediktsson, X.~Zhu, L.~Rittner, and R.~A. Lotufo,
\newblock ``Extinction profiles for the classification of remote sensing
  data,''
\newblock {\em IEEE Trans. Geosci. Remote Sens.}, vol. 54, no. 10, pp.
  5631--5645, 2016.

\bibitem{ghamisi2017lidar}
P.~Ghamisi and B.~Hoefle,
\newblock ``Lidar data classification using extinction profiles and a composite
  kernel support vector machine,''
\newblock {\em IEEE Geosci. Remote Sens. Lett.}, vol. 14, no. 5, pp. 659--663,
  2017.

\bibitem{fang2017extinction}
L.~Fang, N.~He, S.~Li, P.~Ghamisi, and J.~Benediktsson,
\newblock ``Extinction profiles fusion for hyperspectral images
  classification,''
\newblock {\em IEEE Trans. Geosci. Remote Sens.}, vol. 56, no. 3, pp.
  1803--1815, 2017.

\bibitem{xia2018random}
J.~Xia, P.~Ghamisi, N.~Yokoya, and A.~Iwasaki,
\newblock ``Random forest ensembles and extended multiextinction profiles for
  hyperspectral image classification,''
\newblock {\em IEEE Trans. Geosci. Remote Sens.}, vol. 56, no. 1, pp. 202--216,
  2018.

\bibitem{licciardi2012retrieval}
G.~A. Licciardi, A.~Villa, M.~Dalla Mura, L.~Bruzzone, J.~Chanussot, and
  J.~Benediktsson,
\newblock ``Retrieval of the height of buildings from worldview-2 multi-angular
  imagery using attribute filters and geometric invariant moments,''
\newblock {\em IEEE J. Sel. Topics Appl. Earth Observ. Remote Sens.}, vol. 5,
  no. 1, pp. 71--79, 2012.

\bibitem{hu1962visual}
M.~K. Hu,
\newblock ``Visual pattern recognition by moment invariants,''
\newblock {\em IRE trans. inform. theory}, vol. 8, no. 2, pp. 179--187, 1962.

\bibitem{wu2018msri}
X.~Wu, D.~Hong, P.~Ghamisi, W.~Li, and R.~Tao,
\newblock ``Msri-ccf: Multi-scale and rotation-insensitive convolutional
  channel features for geospatial object detection,''
\newblock {\em Remote Sens.}, vol. 10, no. 12, pp. 1990, 2018.

\bibitem{achanta2012slic}
R.~Achanta, A.~Shaji, K.~Smith, A.~Lucchi, P.~Fua, and S.~S{\"u}sstrunk,
\newblock ``S{LIC} superpixels compared to state-of-the-art superpixel
  methods,''
\newblock {\em IEEE Trans. Pattern Anal. Mach. Intell.}, vol. 34, no. 11, pp.
  2274--2282, 2012.

\bibitem{liu2014rotation}
K.~Liu, H.~Skibbe, T.~Schmidt, T.~Blein, K.~Palme, T.~Brox, and O.~Ronneberger,
\newblock ``Rotation-invariant hog descriptors using fourier analysis in polar
  and spherical coordinates,''
\newblock {\em Int. J. Comput. Vis.}, vol. 106, no. 3, pp. 342--364, 2014.

\bibitem{hong2016robust}
D.~Hong, W.~Liu, X.~Wu, Z.~Pan, and J.~Su,
\newblock ``Robust palmprint recognition based on the fast variation
  vese--osher model,''
\newblock {\em Neurocomputing}, vol. 174, pp. 999--1012, 2016.

\bibitem{wu2019orsim}
X.~Wu, D.~Hong, J.~Tian, J.~Chanussot, W.~Li, and R.~Tao,
\newblock ``O{RSIm} {D}etector: A novel object detection framework in optical
  remote sensing imagery using spatial-frequency channel features,''
\newblock {\em IEEE Trans. Geosci. Remote Sens.}, vol. 57, no. 7, pp.
  5146--5158, 2019.

\bibitem{wu2019fourier}
X.~Wu, D.~Hong, J.~Chanussot, Y.~Xu, R.~Tao, and Y.~Wang,
\newblock ``Fourier-based rotation-invariant feature boosting: An efficient
  framework for geospatial object detection,''
\newblock {\em IEEE Geosci. Remote Sens. Lett.}, 2019,
\newblock DOI:10.1109/LGRS.2019.2919755.

\bibitem{reisert2008equivariant}
M.~Reisert and H.~Burkhardt,
\newblock ``Equivariant holomorphic filters for contour denoising and rapid
  object detection,''
\newblock {\em IEEE Trans. Image Process.}, vol. 17, no. 2, pp. 190--203, 2008.

\bibitem{wang2009rotational}
Q.~Wang, O.~Ronneberger, and H.~Burkhardt,
\newblock ``Rotational invariance based on fourier analysis in polar and
  spherical coordinates,''
\newblock {\em IEEE Trans. Pattern Anal. Mach. Intell.}, vol. 31, no. 9, pp.
  1715--1722, 2009.

\bibitem{vedaldi2011learning}
A.~Vedaldi, M.~Blaschko, and A.~Zisserman,
\newblock ``Learning equivariant structured output svm regressors,''
\newblock in {\em Proc. ICCV}. IEEE, 2011, pp. 959--966.

\bibitem{jacovitti2000multiresolution}
G.~Jacovitti and A.~Neri,
\newblock ``Multiresolution circular harmonic decomposition,''
\newblock {\em IEEE Trans. Signal Process.}, vol. 48, no. 11, pp. 3242--3247,
  2000.

\bibitem{giannakis1989signal}
G.~B. Giannakis,
\newblock ``Signal reconstruction from multiple correlations: frequency-and
  time-domain approaches,''
\newblock {\em J. Opt. Soc. Am. A.}, vol. 6, no. 5, pp. 682--697, Feb 1989.

\bibitem{hong2017learning}
D.~Hong, N.~Yokoya, and X.~Zhu,
\newblock ``Learning a robust local manifold representation for hyperspectral
  dimensionality reduction,''
\newblock {\em IEEE J. Sel. Topics Appl. Earth Observ. Remote Sens.}, vol. 10,
  no. 6, pp. 2960--2975, 2017.

\bibitem{dollar2014fast}
P.~Doll{\'a}r, R.~Appel, S.~Belongie, and P.~Perona,
\newblock ``Fast feature pyramids for object detection,''
\newblock {\em IEEE Trans. Pattern Anal. Mach. Intell.}, vol. 36, no. 8, pp.
  1532--1545, 2014.

\bibitem{marpu2012classification}
P.~R. Marpu, M.~Pedergnana, M.~D. Mura, S.~Peeters, J.~Benediktsson, and
  L.~Bruzzone,
\newblock ``Classification of hyperspectral data using extended attribute
  profiles based on supervised and unsupervised feature extraction
  techniques,''
\newblock {\em Int. J. Image Data Fusion}, vol. 3, no. 3, pp. 269--298, 2012.

\bibitem{hong2018joint}
D.~Hong, N.~Yokoya, J.~Xu, and X.~Zhu,
\newblock ``Joint \& progressive learning from high-dimensional data for
  multi-label classification,''
\newblock in {\em Proc. ECCV}, 2018, pp. 469--484.

\bibitem{hong2019learning}
D.~Hong, N.~Yokoya, J.~Chanussot, J.~Xu, and X.~X. Zhu,
\newblock ``Learning to propagate labels on graphs: An iterative multitask
  regression framework for semi-supervised hyperspectral dimensionality
  reduction,''
\newblock {\em ISPRS J. Photogramm. Remote Sens.}, vol. 158, pp. 35--49, 2019.

\bibitem{rasti2017hyperspectral}
B.~Rasti, P.~Ghamisi, and R.~Gloaguen,
\newblock ``Hyperspectral and lidar fusion using extinction profiles and total
  variation component analysis,''
\newblock {\em IEEE Trans. Geosci. Remote Sens.}, vol. 55, no. 7, pp.
  3997--4007, 2017.

\end{thebibliography}

\begin{IEEEbiography}[{\includegraphics[width=1in,height=1.25in,clip,keepaspectratio]{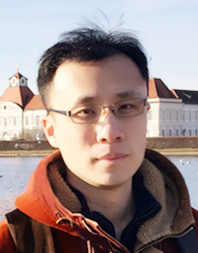}}]{Danfeng Hong}
(S'16-M'19) was born Shandong Province, China, in 1989. He received the B.Sc. degree in computer science and technology, Neusoft College of Information, Northeastern University, China, in 2012, the M.Sc. degree (summa cum laude) in computer vision, College of Information Engineering, University, China, in 2015, the Dr. -Ing degree (summa cum laude) in Signal Processing in Earth Observation (SiPEO), Technical University of Munich (TUM), Munich, Germany, in 2019.

Since 2015, he also worked as a Research Associate at the Remote Sensing Technology Institute (IMF), German Aerospace Center (DLR), Oberpfaffenhofen, Germany. Currently, he is research scientist and leads a Spectral Vision group at EO Data Science, IMF, DLR. In 2018 and 2019, he was a visiting scholar in GIPSA-lab, Grenoble INP, CNRS, Univ. Grenoble Alpes, Grenoble, France, and in RIKEN Artificial Intelligent Project (AIP), RIKEN, Tokyo, Japan.

His research interests include signal / image processing and analysis, pattern recognition, machine / deep learning and their applications in Earth Vision.
\end{IEEEbiography}
\vskip -2\baselineskip plus -1fil
\begin{IEEEbiography}[{\includegraphics[width=1in,height=1.25in,clip,keepaspectratio]{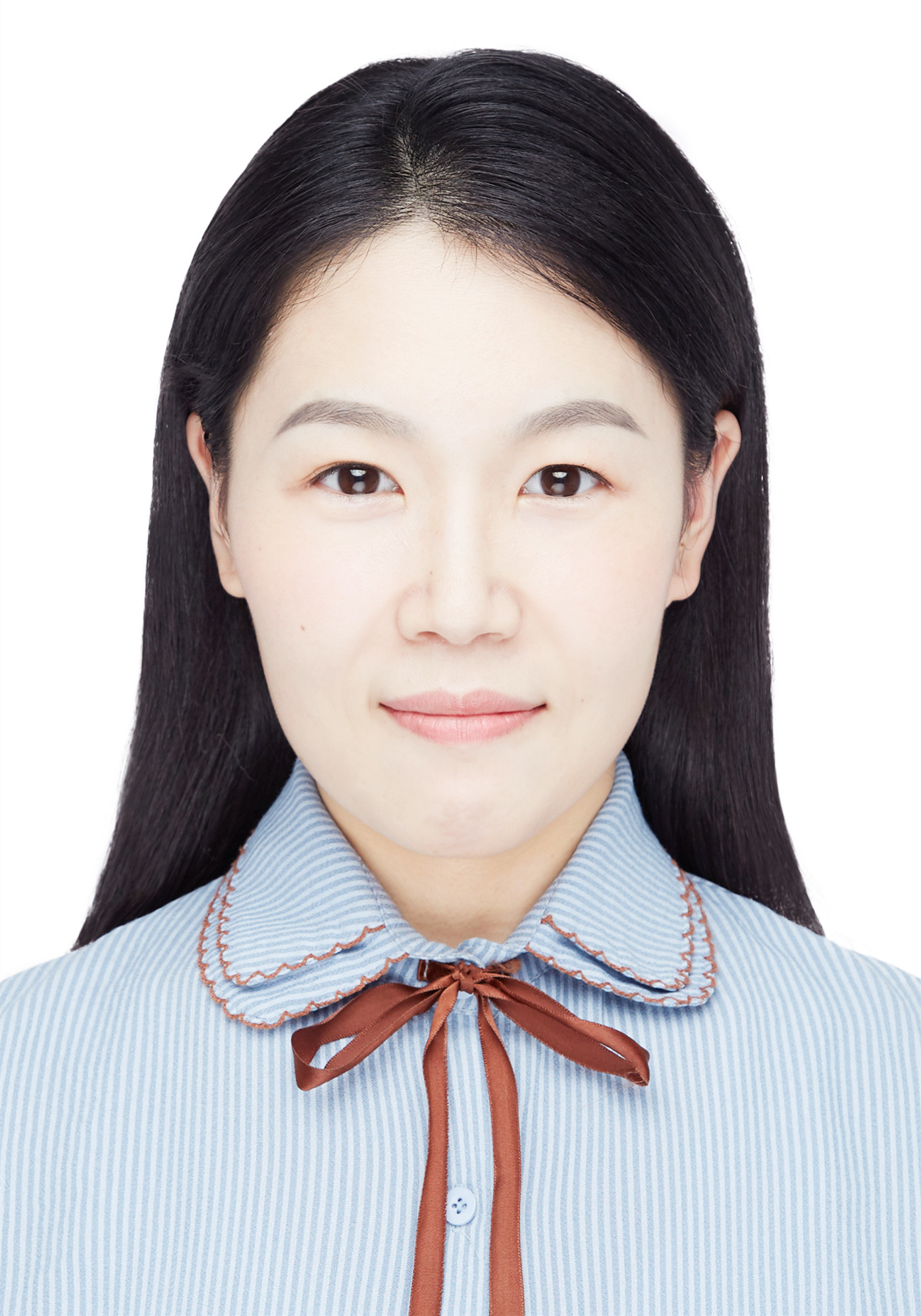}}]{Xin Wu}
(S'19) received the B.S. degree in Science and Technology of Electronic Information department from Mudanjiang Normal University, China, in 2011. She received the M.Sc. degree in Computer Science and Technology, Qingdao University, China, in 2014. She is currently pursuing her Ph.D. degree with Information and Communication Engineering in Beijing Institute of Technology (BIT), Beijing, China, since September 2014.

In 2018, she was a visiting student at the Photogrammetry and Image Analysis department of the Remote Sensing Technology Institute (IMF), German Aerospace Center (DLR), Oberpfaffenhofen, Germany.

Her research interests include signal / image processing, fractional fourier transform, deep learning and their applications in biometrics and geospatial object detection.
\end{IEEEbiography}
\vskip -2\baselineskip plus -1fil
\begin{IEEEbiography}[{\includegraphics[width=1in,height=1.25in,clip,keepaspectratio]{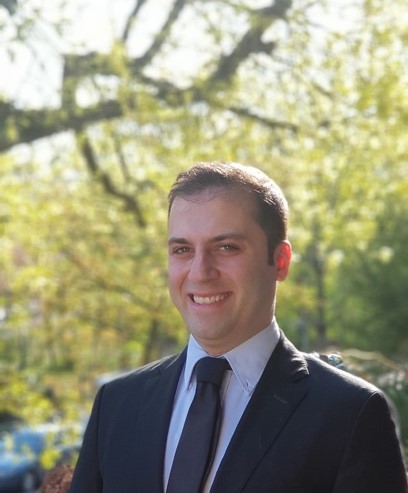}}]{Pedram Ghamisi} (S'12-M'15-SM'17) received the B.Sc. degree in civil (survey) engineering from the Tehran South Campus of Azad University, Tehran, Iran, in 2008, the M.Sc. degree (Hons.) in remote sensing with  the K. N. Toosi University of Technology, Tehran, Iran, in 2012, and the Ph.D. degree in electrical and computer engineering with the University of Iceland, Reykjavik, Iceland, in 2015.
In 2013 and 2014, he was with the School of Geography, Planning and Environmental Management, University of Queensland, Brisbane, QLD, Australia. In 2015, he was a Alexander von Humboldt Research Fellow with the Technical University of Munich (TUM), Munich, Germany, and Heidelberg University, Heidelberg, Germany. From 2015 to 2018, he was a Research Scientist with the German Aerospace Center (DLR), Oberpfaffenhofen, Germany. Since 2018, he has been working as the Head of the Machine Learning Group, Helmholtz-Zentrum Dresden-Rossendorf (HZDR). He is also the CTO and the Co-Founder of VasoGnosis Inc., Milwaukee, WI, USA, where he is involved in the development of advanced diagnostic and analysis tools for brain diseases using cloud computing and deep learning algorithms. He is also the vice-chair of the IEEE Image Analysis and Data Fusion Committee. His research interests include interdisciplinary research on remote sensing and machine (deep) learning, image and signal processing, and multisensor data fusion.
Dr. Ghamisi was a recipient of the Best Researcher Award for M.Sc. students at the K. N. Toosi University of Technology in the academic year of 2010-2011, the IEEE Mikio Takagi Prize for winning the Student Paper Competition at IEEE International Geoscience and Remote Sensing Symposium (IGARSS) in 2013, the Talented International Researcher by Iran’s National Elites Foundation in 2016, the first prize of the data fusion contest organized by the Image Analysis and Data Fusion Technical Committee (IADF) of IEEE-GRSS in 2017, the Best Reviewer Prize of IEEE Geoscience and Remote Sensing Letters (GRSL) in 2017, the Alexander von Humboldt Fellowship from the Technical University of Munich, and the High Potential Program Award from HZDR. He serves as an Associate Editor for MDPI-Remote Sensing and IEEE GRSL.

\end{IEEEbiography}
\vskip -2\baselineskip plus -1fil
\begin{IEEEbiography}[{\includegraphics[width=1in,height=1.25in,clip,keepaspectratio]{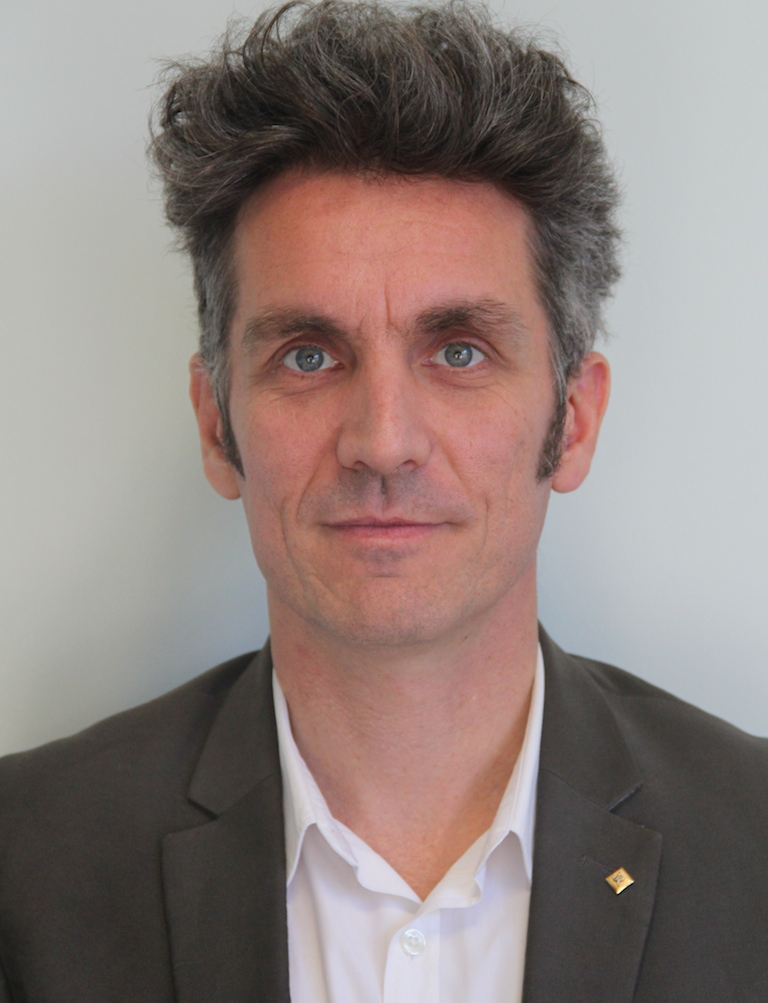}}]{Jocelyn Chanussot}
(M'04–SM'04-F'12) received the M.Sc. degree in electrical engineering from the Grenoble Institute of Technology (Grenoble INP), Grenoble, France, in 1995, and the Ph.D. degree from the Université de Savoie, Annecy, France, in 1998. In 1999, he was with the Geography Imagery Perception Laboratory for the Delegation Generale de l'Armement (DGA - French National Defense Department). Since 1999, he has been with Grenoble INP, where he is currently a Professor of signal and image processing. He is conducting his research at GIPSA-Lab. His research interests include image analysis, multicomponent image processing, nonlinear filtering, and data fusion in remote sensing. He has been a visiting scholar at Stanford University (USA), KTH (Sweden) and NUS (Singapore). Since 2013, he is an Adjunct Professor of the University of Iceland. In 2015-2017, he was a visiting professor at the University of California, Los Angeles (UCLA).

Dr. Chanussot is the founding President of IEEE Geoscience and Remote Sensing French chapter (2007-2010) which received the 2010 IEEE GRSS Chapter Excellence Award. He was the co-recipient of the NORSIG 2006 Best Student Paper Award, the IEEE GRSS 2011 and 2015 Symposium Best Paper Award, the IEEE GRSS 2012 Transactions Prize Paper Award and the IEEE GRSS 2013 Highest Impact Paper Award. He was a member of the IEEE Geoscience and Remote Sensing Society AdCom (2009-2010), in charge of membership development. He was the General Chair of the first IEEE GRSS Workshop on Hyperspectral Image and Signal Processing, Evolution in Remote sensing (WHISPERS). He was the Chair (2009-2011) and Co-chair of the GRS Data Fusion Technical Committee (2005-2008). He was a member of the Machine Learning for Signal Processing Technical Committee of the IEEE Signal Processing Society (2006-2008) and the Program Chair of the IEEE International Workshop on Machine Learning for Signal Processing (2009). He was an Associate Editor for the IEEE Geoscience and Remote Sensing Letters (2005-2007) and for Pattern Recognition (2006-2008). He was the Editor-in-Chief of the IEEE Journal of Selected Topics in Applied Earth Observations and Remote Sensing (2011-2015). Since 2007, he is an Associate Editor for the IEEE Transactions on Geoscience and Remote Sensing, and since 2018, he is also an Associate Editor for the IEEE Transactions on Image Processing. In 2013, he was a Guest Editor for the Proceedings of the IEEE and in 2014 a Guest Editor for the IEEE Signal Processing Magazine. He is a Fellow of the IEEE and a member of the Institut Universitaire de France (2012-2017).
\end{IEEEbiography}
\vskip -2\baselineskip plus -1fil
\begin{IEEEbiography}[{\includegraphics[width=1in,height=1.25in,clip,keepaspectratio]{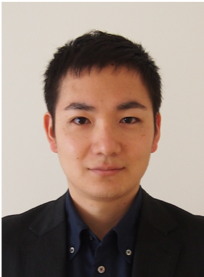}}]{Naoto Yokoya}
Naoto Yokoya (S'10-M'13) received the M.Eng. and Ph.D. degrees in aerospace engineering from the University of Tokyo, Tokyo, Japan, in 2010 and 2013, respectively.

He is currently a Unit Leader at the RIKEN Center for Advanced Intelligence Project, Tokyo, Japan, where he leads the Geoinformatics Unit since 2018. He is also a visiting Associate Professor at Tokyo University of Agriculture and Technology since 2019. He was an Assistant Professor at the University of Tokyo from 2013 to 2017. In 2015-2017, he was an Alexander von Humboldt Fellow, working at the German Aerospace Center (DLR), Oberpfaffenhofen, and Technical University of Munich (TUM), Munich, Germany. His research is focused on the development of image processing, data fusion, and machine learning algorithms for understanding remote sensing images, with applications to disaster management.

Dr. Yokoya won the first place in the 2017 IEEE Geoscience and Remote Sensing Society (GRSS) Data Fusion Contest organized by the Image Analysis and Data Fusion Technical Committee (IADF TC). He is the Chair (2019-2021) and was a Co-Chair (2017-2019) of IEEE GRSS IADF TC and also the secretary of IEEE GRSS All Japan Joint Chapter since 2018. He is an Associate Editor for the IEEE Journal of Selected Topics in Applied Earth Observations and Remote Sensing (JSTARS) since 2018. He is/was a Guest Editor for the IEEE JSTARS in 2015-2016, for Remote Sensing in 2016-2019, and for the IEEE Geoscience and Remote Sensing Letters (GRSL) in 2018-2019.
\end{IEEEbiography}
\begin{IEEEbiography}[{\includegraphics[width=1in,height=1.25in,clip,keepaspectratio]{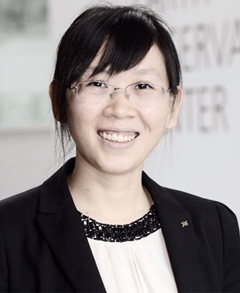}}]{Xiao Xiang Zhu}(S'10--M'12--SM'14) received the Master (M.Sc.) degree, her doctor of engineering (Dr.-Ing.) degree and her “Habilitation” in the field of signal processing from Technical University of Munich (TUM), Munich, Germany, in 2008, 2011 and 2013, respectively.

She is currently the Professor for Signal Processing in Earth Observation (www.sipeo.bgu.tum.de) at Technical University of Munich (TUM) and German Aerospace Center (DLR); the head of the department ``EO Data Science'' at DLR's Earth Observation Center; and the head of the Helmholtz Young Investigator Group ``SiPEO'' at DLR and TUM. Prof. Zhu was a guest scientist or visiting professor at the Italian National Research Council (CNR-IREA), Naples, Italy, Fudan University, Shanghai, China, the University  of Tokyo, Tokyo, Japan and University of California, Los Angeles, United States in 2009, 2014, 2015 and 2016, respectively. Her main research interests are remote sensing and Earth observation, signal processing, machine learning and data science, with a special application focus on global urban mapping.

Dr. Zhu is a member of young academy (Junge Akademie/Junges Kolleg) at the Berlin-Brandenburg Academy of Sciences and Humanities and the German National  Academy of Sciences Leopoldina and the Bavarian Academy of Sciences and Humanities. She is an associate Editor of IEEE Transactions on Geoscience and Remote Sensing.
\end{IEEEbiography}
\end{document}